\documentclass{article}
\usepackage{amssymb}
\usepackage{graphicx}

\usepackage{booktabs}  
\usepackage{longtable}

\usepackage{hyperref}
\usepackage{url}
\usepackage[nolist]{acronym} 
\usepackage{xspace} 
\usepackage{wrapfig}

\usepackage{amsmath,amsfonts,bm}

\usepackage[title]{appendix}

\newcommand{\pose}[1]{\boldsymbol{T}_{#1}}

\usepackage[final]{corl_2025} 

\title{Learning a Thousand Tasks in a Day}

\author{
	Kamil Dreczkowski $^{1\dagger\ast}$,
	Pietro Vitiello$^{1\dagger\ast}$,
	Vitalis Vosylius$^{1}$,
        Edward Johns$^{1}$\and
	\small$^{1}$The Robot Learning Lab at Imperial College London, London, SW7 2AZ, UK.\and
	\small$^\ast$Corresponding authors. Email: kamil-dreczkowski@outlook.com, pv2017@ic.ac.uk\and
	\small$^\dagger$These authors contributed equally to this work.
}

\begin{document}
\maketitle


\let\oldthefootnote\thefootnote
\renewcommand{\thefootnote}{\relax}
\footnotetext{This is the author's version of the work. It is posted here by permission of the AAAS for personal use, not for redistribution. The definitive version was published in Science Robotics on 12 November 2025, DOI: \url{https://www.science.org/doi/10.1126/scirobotics.adv7594}.}
\renewcommand{\thefootnote}{\oldthefootnote}
\let\oldthefootnote\undefined

\begin{abstract}
Humans are remarkably efficient at learning tasks from demonstrations, but today's imitation learning methods for robot manipulation often require hundreds or thousands of demonstrations per task. We investigate two fundamental priors for improving learning efficiency: decomposing manipulation trajectories into sequential alignment and interaction phases, and retrieval-based generalisation. Through 3,450 real-world rollouts, we systematically study this decomposition. We compare different design choices for the alignment and interaction phases, and examine generalisation and scaling trends relative to today's dominant paradigm of behavioural cloning with a single-phase monolithic policy. In the few-demonstrations-per-task regime ($<$10 demonstrations), decomposition achieves an order of magnitude improvement in data efficiency over single-phase learning, with retrieval consistently outperforming behavioural cloning for both alignment and interaction. Building on these insights, we develop Multi-Task Trajectory Transfer (MT3), an imitation learning method based on decomposition and retrieval. MT3 learns everyday manipulation tasks from as little as a single demonstration each, whilst also generalising to novel object instances. This efficiency enables us to teach a robot 1,000 distinct everyday tasks in under 24 hours of human demonstrator time. Through 2,200 additional real-world rollouts, we reveal MT3's capabilities and limitations across different task families. Videos of our experiments can be found on our \href{https://www.robot-learning.uk/learning-1000-tasks}{\color{blue}website}.
\end{abstract}


\begin{figure}[t!]
    \centering
    \includegraphics[width=\linewidth]{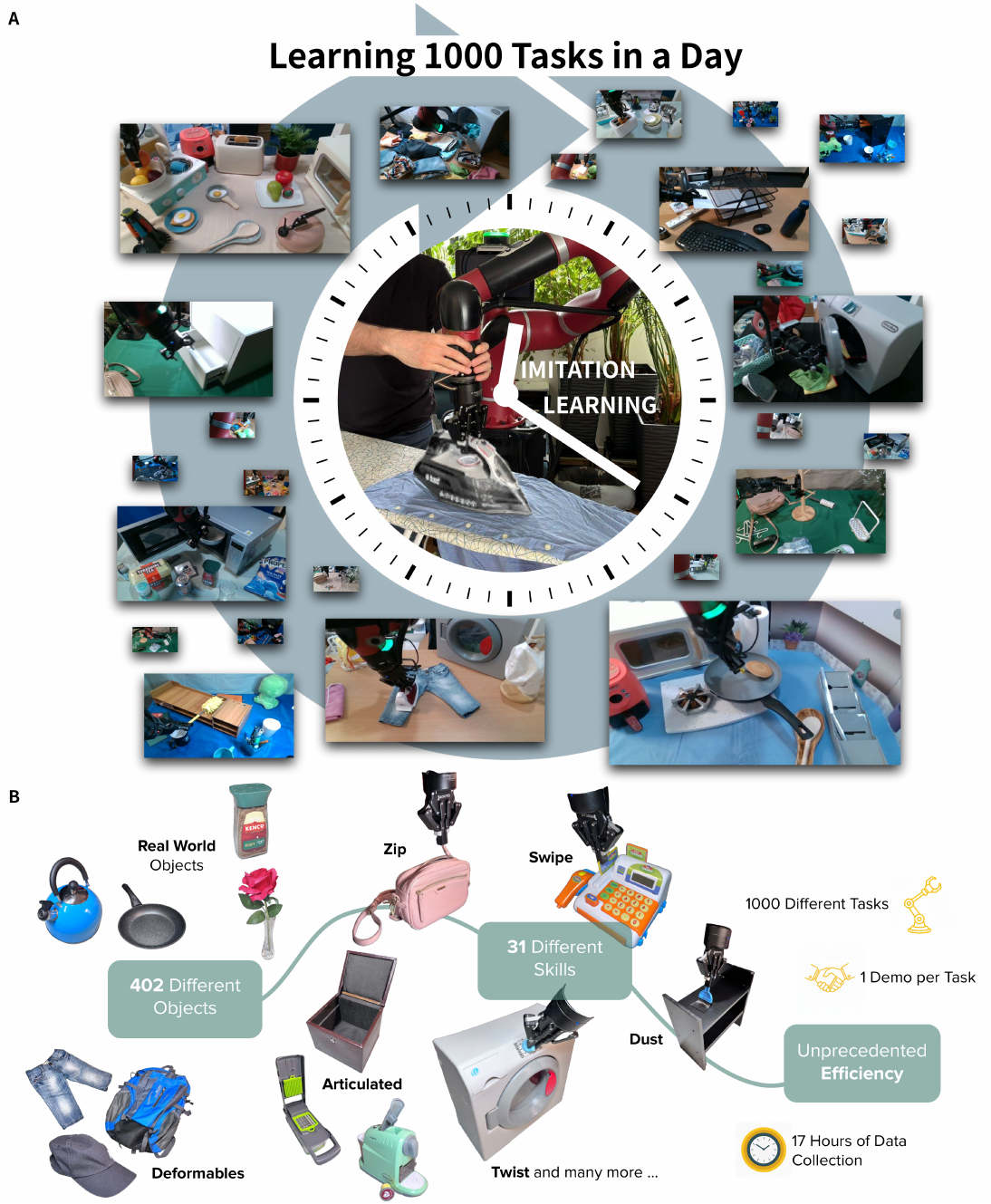}
    \caption{\textbf{Learning a thousand tasks in a day.} (\textbf{A}) Illustration of 1,000 tasks taught in less than a day. The arrow represents the passing of time, whereas each image is a frame from a real-world rollout of one of the tasks. (\textbf{B}) Illustration of some information regarding the 1,000 tasks dataset. We provide examples of some objects used and some of the skills evaluated.}
    \label{fig:fig1}
\end{figure}


\section{Introduction}
	
Humans and animals demonstrate remarkable learning efficiency through imitation. Infants learn manipulation skills substantially faster when guided by expert demonstrations~\cite{SOMOGYI2015186, Fagard2016-st}, primates learn manipulation tasks from fewer than five demonstrations~\cite{Hayes1952, Horner2004-ay, Call2004-pg, Rigamonti2005-wd, Tennie2006}, and rodents acquire both behaviour and navigation skills from fewer than ten demonstrations~\cite{MEISTER2022102555}. In stark contrast, robots lag far behind their biological counterparts, often requiring hundreds or thousands of demonstrations per task~\cite{jang2021bcz, Jiang2022VIMA, Shafiullah2022, Brohan2023rt1, rt2, Zhao2023aloha, bharadhwaj2023roboagent, Kim2024, black2024pi0visionlanguageactionflowmodel, octo_2023}.

State-of-the-art imitation learning systems using Behavioural Cloning (BC) exemplify this inefficiency. Behavior Cloning with Zero-Shot Task Generalization (BC-Z) required $\sim$26K demonstrations for 100 tasks~\cite{jang2021bcz}, Robotics Transformer 1 (RT-1) needed $\sim$130K demonstrations across 744 tasks ~\cite{Brohan2023rt1}, and Multi-Task Action-Chunking Transformers (MT-ACT) collected 7.5K demonstrations for 38 tasks~\cite{bharadhwaj2023roboagent}— all averaging 175-250 demonstrations per task. For complex bimanual manipulation, A Low-cost Open-source Hardware System for Bimanual Teleoperation (ALOHA) Unleashed~\cite{zhao2024aloha} suggests the need for $\sim$8K demonstrations per task.

Although these methods can be effective at scale, scaling to thousands of tasks would require massive real-world datasets demanding enormous financial and human resources. Improving learning efficiency is thus crucial for reducing the data requirements for highly capable and general robotic systems. To this end, we study two priors for more data-efficient imitation learning: trajectory decomposition and retrieval-based generalisation.

The first prior decomposes manipulation trajectories into alignment and interaction phases. In contrast to standard BC approaches that learn manipulation with a single monolithic policy, decomposition-based methods deploy two independent policies sequentially. Firstly, an alignment policy positions the robot's end-effector, or a grasped object, relative to the target object. Secondly, an interaction policy performs the object manipulation. For alignment, past research has explored using pose estimation~\cite{lee2020guapo, vitiello2023one} and visual servoing~\cite{johns2021coarse-to-fine, valassakis2021coarse-to-fine, dipalo2021learning, valassakis2022dome, zhao2024alearningbased, dipalo2024dinobot, wang2025one-shot}. For interaction, prior work has primarily focused on reinforcement learning~\cite{lee2020guapo, zhao2024alearningbased} and open-loop replay~\cite{johns2021coarse-to-fine, valassakis2021coarse-to-fine, dipalo2021learning, valassakis2022dome, vitiello2023one, dipalo2024dinobot, wang2025one-shot}. Unlike past work that typically focuses on single-task learning with one specific alignment and interaction policy combination, we systematically compare four different combinations (pose estimation vs. BC for alignment; open-loop replay vs. BC for interaction) when learning multiple tasks.

The second prior uses retrieval-based generalisation as an alternative to BC. Recent retrieval methods for manipulation include Flow-Guided Data Retrieval for Few-Shot Imitation Learning (FlowRetrieval)~\cite{lin2024flowretrieval}, Skill-Augmented Imitation Learning with prior Retrieval (SAILOR)~\cite{nasiriany2023learning}, and Behavior Retrieval~\cite{Du2023behavior}, which leverage optical flow matching, latent skill spaces, and task-specific querying, respectively. Whereas these approaches primarily retrieve data prior to policy training, our methods retrieve demonstrations at test time.

After retrieving the most appropriate demonstration using language-and-geometry-based matching, we perform alignment through pose estimation~\cite{lee2020guapo, vitiello2023one} and execute interaction via open-loop replay~\cite{johns2021coarse-to-fine, valassakis2021coarse-to-fine, dipalo2021learning, valassakis2022dome, vitiello2023one, dipalo2024dinobot, wang2025one-shot}. Unlike Visual Imitation through Nearest Neighbors (VINN)~\cite{pari2021surprising} which uses RGB-based retrieval to identify and obtain actions from image-action pairs throughout task execution, we use language and geometry to retrieve a complete trajectory prior to task execution.

Through 3,450 real-world rollouts across 70 objects, we systematically analyse the effects of trajectory decomposition and retrieval-based generalisation on learning efficiency. Although existing research typically focuses on learning with abundant per-task demonstrations, we focus on the more practical scenario where demonstrations are limited —  a critical gap in current literature. We study all four combinations of BC and retrieval-based policies when used for alignment and interaction, and compare them against a standard monolithic BC method that learns entire manipulation trajectories without decomposition. 

Our results reveal key insights into the trade-offs between the number of demonstrations per task, the number of tasks and object instances being learned, and the task performance. In the very low-data-per-task regime ($<10$ demonstrations per task), decomposition yields an order of magnitude improvement in data efficiency compared to learning trajectories with a single monolithic policy. Furthermore, retrieval-based methods prove more effective for generalisation, consistently outperforming BC alternatives across both alignment and interaction phases. However, as demonstrations become more abundant or task diversity increases (distributing a fixed demonstration budget across more tasks), monolithic BC exhibits better scaling trends.

Building on these insights, we develop Multi-Task Trajectory Transfer (MT3), a fully retrieval-based decomposition method that leverages retrieval for both alignment and interaction. Although MT3 demonstrates particularly strong performance in our controlled experiments, a key question remains: is MT3 a viable approach for learning a very large number of tasks from minimal per-task data? To answer this question, we conduct a large-scale evaluation in terms of task and object diversity: teaching a robot 1,000 distinct everyday tasks — involving interactions with over 400 objects — from single demonstrations in less than 24 hours (see Figure~\ref{fig:fig1} and our \href{https://www.robot-learning.uk/learning-1000-tasks}{\color{blue}website} for a video showing us collecting all 1,000 demonstrations). Through 2,200 experimental rollouts, we find that MT3 is indeed capable of scaling to very large numbers of tasks, and gain insights into its limitations and failure modes.

In summary, our work makes three key contributions. Firstly, we provide a systematic evaluation of multi-task imitation learning in the few-demonstration-per-task regime, addressing a critical gap in current literature. Secondly, we introduce MT3 and demonstrate that retrieval-based decomposition methods offer an attractive alternative to monolithic BC when demonstration data is limited. Thirdly, we validate these findings at scale by learning 1,000 distinct manipulation tasks from a single demonstration each, challenging the assumption that complex neural policies are necessary for large-scale robot learning, whilst gaining insights into MT3's limitations and failure modes.

\section{Results}

\subsection{Experimental setup}

\begin{figure}[h]
    \centering
    \includegraphics[width=0.85\linewidth]{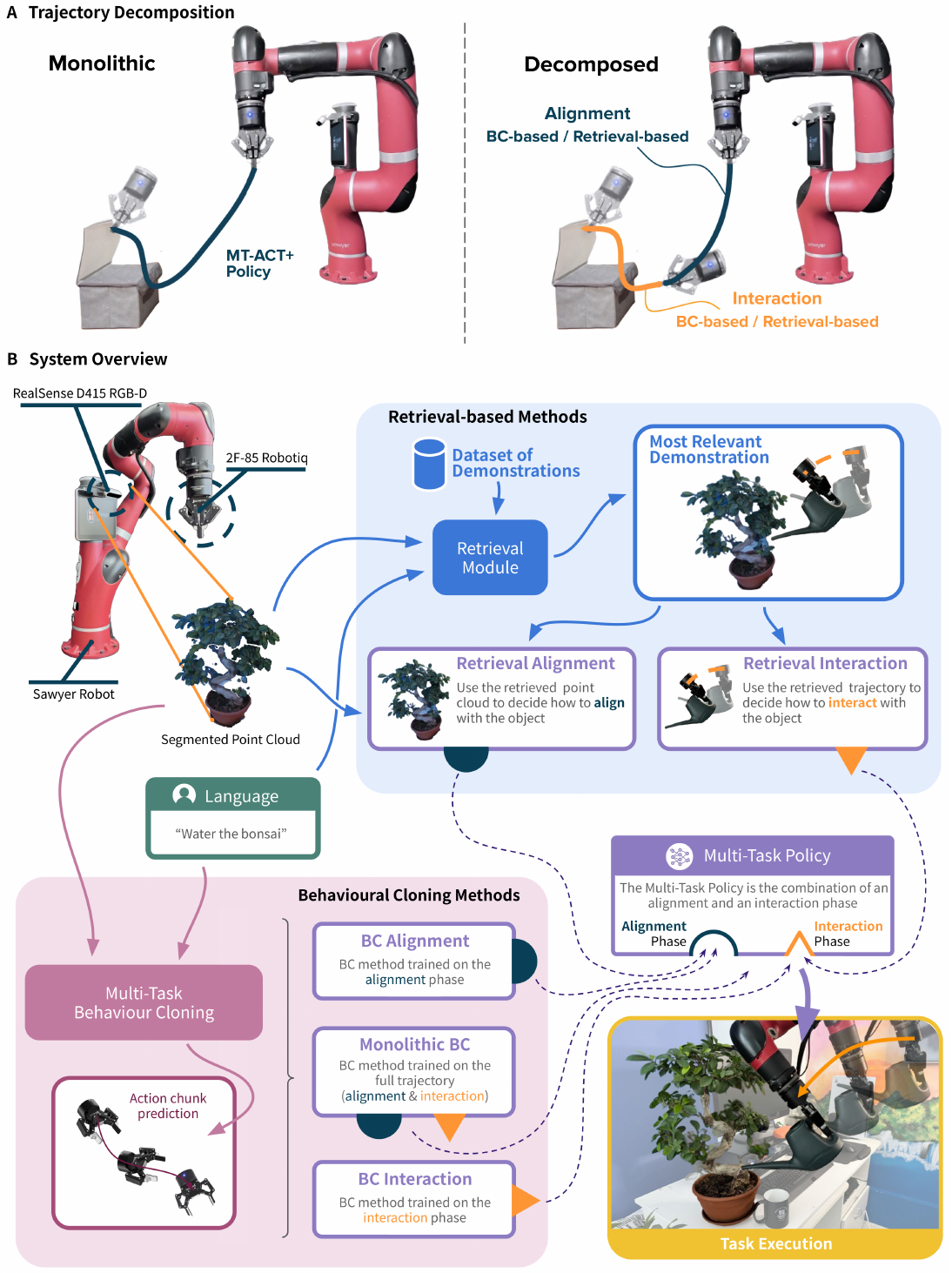}
    \caption{\textbf{Trajectory decomposition and overview of policy designs.} (\textbf{A}) Trajectories are decomposed into alignment and interaction phases. Monolithic approaches use a single policy for entire trajectories. Decomposition-based approaches use two specialised policies: one for end-effector alignment with target objects, and another for precise manipulations. We explore both BC and retrieval-based methods for each phase of this decomposition. (\textbf{B}) A multi-task policy (purple) processes a segmented point cloud and task description as input and outputs robot actions. This can either be a monolithic policy or the combination of an alignment and an interaction policies. Retrieval-based policies (blue) use a retrieved demonstration as context to guide execution. Behavioural Cloning policies (pink) directly predict actions through a neural network.}
    \label{fig:fig2}
\end{figure}

We focused on teaching a robot multiple tasks, where each task involved a single interaction between the robot's end-effector, or a grasped object, and a target object. For tasks involving grasped objects, we assumed that their pose in the gripper was the same during demonstrations and testing. This formulation covers most common manipulation tasks — from grasping to insertion to tool usage. Although we focused on single-interaction tasks, multi-step behaviours such as pick-and-place operations could be achieved by chaining such tasks together using existing high-level planners (see our \href{https://www.robot-learning.uk/learning-1000-tasks}{\color{blue}website} for videos). 

Our evaluation considered both seen tasks and unseen tasks where methods had to generalise to novel objects. For clarity, we defined three terms. A macro skill is a broad manipulation primitive defined by its core interaction type, for example, ``open", ``insert", or ``fold". A micro skill is a macro skill specialised for a specific object category that requires a distinct motion profile. Different motion profiles for the same object category constitute different micro skills, including ``open oven door sideways" versus ``open oven door downward". A task is a micro skill executed on a specific object instance, such as, ``unzip the round pink handbag”. 

Our experimental hardware consisted of a Sawyer robot equipped with a 2F-85 Robotiq gripper (Figure~\ref{fig:fig2}). For perception, we used a single RealSense D415 RGB-D camera mounted on the robot's head. To ensure a fair comparison, we established a consistent system architecture across all methods. The robot receives two inputs: a segmented point cloud of the target object and a language description of the task. A multi-task policy processes these inputs to generate robot actions. In terms of policy design, we compared four decomposition-based methods against a monolithic BC baseline that learned entire trajectories.

\subsection{Policy designs}
\subsubsection{\textit{Decomposition-based methods}}

Decomposition-based methods divide manipulation trajectories into two phases (Figure~\ref{fig:fig2}.A). The alignment phase involves moving the robot's end-effector to a pose suitable for the subsequent manipulation, where only the final positioning matters, not the specific path taken. For example, positioning a plug in front of a socket can be achieved through many different trajectories. The interaction phase involves the actual manipulation, requiring precise trajectory execution. For example, the plug insertion motion must be carefully controlled to ensure proper connection.

All four decomposition-based methods used two specialised policies: one for alignment and one for interaction. We showed that this specialisation led to efficiency gains compared to using a single policy when learning from limited per-task demonstrations. For each phase, we investigated two alternative approaches: BC and retrieval-based methods. 

BC is a prominent robot manipulation approach that trains neural networks to encode demonstrated behaviours into their weights. We therefore explored applying this technique within the decomposition framework. During training, BC uses all demonstrations to learn a representation that enables generalisation across spatial configurations and object instances based on pose and geometric similarity.

For our BC implementation, we used a transformer-based backbone that employs variational inference \cite{shankar2020, graves2011}, as this architecture has demonstrated efficient learning of various manipulation tasks \cite{bharadhwaj2023roboagent}. Specifically, we adapted the MT-ACT architecture~\cite{bharadhwaj2023roboagent} to handle point cloud inputs and language descriptions. By training this architecture separately on alignment and interaction demonstrations, we obtained specialised BC policies for each phase (Methods subsection ``Behavioural cloning implementation").

An alternative to BC are retrieval-based methods, which fundamentally differ from the former by utilising demonstrations at test time rather than during training. These policies were designed to replicate the behaviour provided to them as context. Therefore, they store all demonstrations in memory, and at inference, retrieve the most relevant one to use as guidance.

Both alignment and interaction retrieval-based policies share a common retrieval step that occurs prior to task execution. This identifies a demonstration of the same manipulation on an object with similar appearance and pose to the test object (Figure \ref{fig:fig2}.B). In our implementation, retrieval leveraged language processing of task descriptions combined with geometry similarity in a learnt latent space. The geometry was extracted from an RGB-D image of the scene before execution. Therefore, when retrieval was used for both alignment and interaction, retrieval was performed once.

After retrieval, the alignment retrieval-based policy uses pose estimation to map the demonstrated alignment pose to the test scene and reaches this pose using motion planning~\cite{vitiello2023one}. The interaction retrieval-based policy executes the retrieved trajectory by replaying demonstrated end-effector velocities in the end-effector frame~\cite{johns2021coarse-to-fine, valassakis2021coarse-to-fine, dipalo2021learning, valassakis2022dome, vitiello2023one, dipalo2024dinobot, wang2025one-shot}. See Methods subsection ``Retrieval-based alignment and interaction" for more details.

For generalisation, retrieval-based policies identify the closest demonstration object and proceed as though the novel object were identical to the training instance. For alignment, they position the robot relative to the novel object as they would for the training one, whereas for interaction, they execute the precise demonstrated trajectory. This approach was effective because optimal trajectories often maintain similar structures across object instances within a category, with task tolerance accommodating geometric variations. For example, when grasping different mugs, although sizes and handle shapes vary, the core grasp motion remains consistent.

Combining BC and retrieval-based policies across both phases (alignment and interaction) created four distinct methods (Figure \ref{fig:fig2}.B). \textbf{BC}-\textbf{BC} uses BC for both alignment and interaction. \textbf{BC}-\textbf{Ret} combines BC alignment with retrieval-based interaction. \textbf{Ret}-\textbf{BC} uses retrieval-based alignment followed by BC interaction. Finally, \textbf{Ret}-\textbf{Ret} employs retrieval-based policies for both alignment and interaction. Throughout this paper, we referred to Ret-Ret as Multi-Task Trajectory Transfer (MT3), since it can be seen as an extension of Trajectory Transfer~\cite{vitiello2023one, schulman2016learning} to the multi-task learning setting.

\begin{figure}[t]
    \centering
    \includegraphics[width=\linewidth]{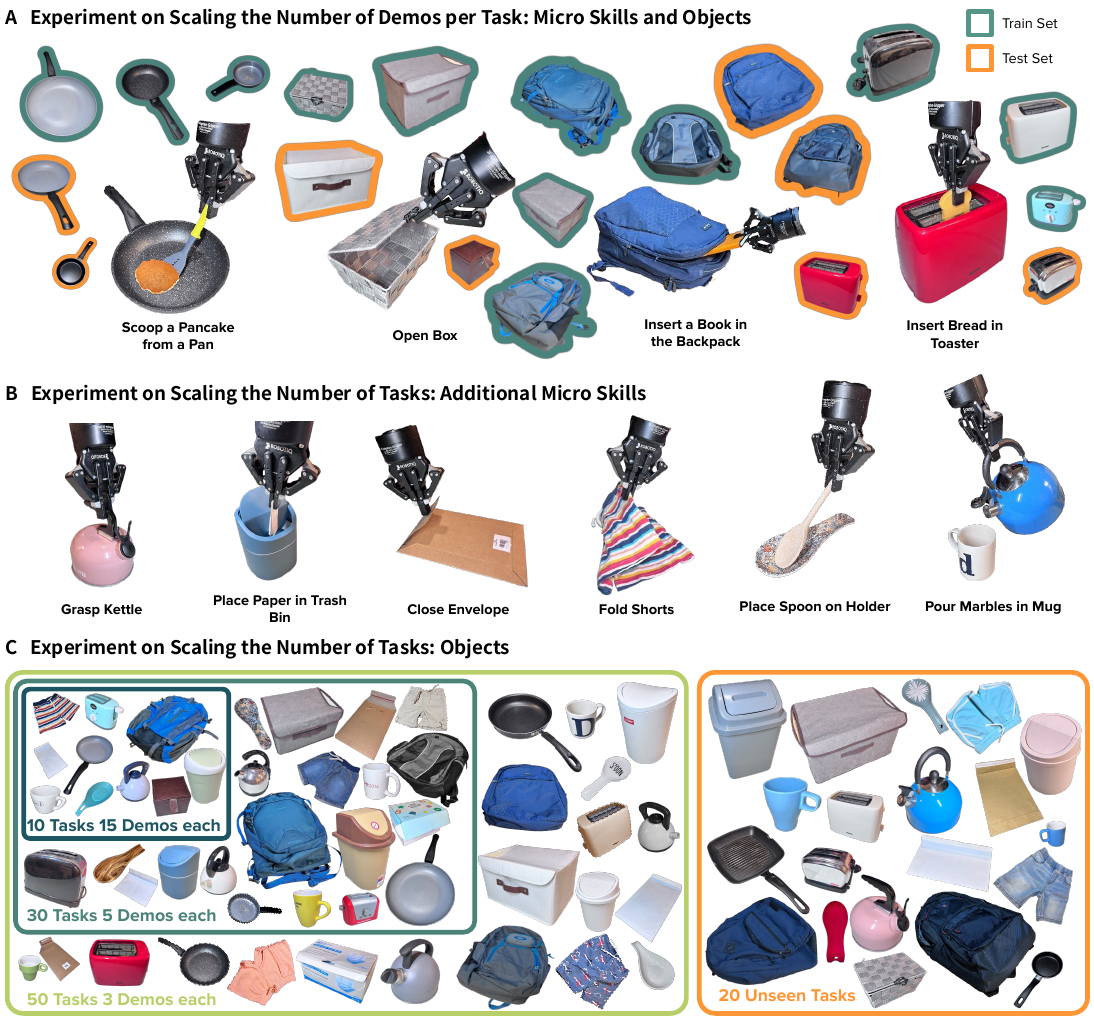}
    \caption{\textbf{Micro skills and objects considered in the scaling experiments.} (\textbf{A}) The micro skills used to evaluate the methods’ response to scaling the demonstrations per task. We also show the various seen and unseen objects used. (\textbf{B}) The micro skills used to evaluate the methods’ response to scaling the number of tasks. These are in addition to those found in (\textbf{A}). (\textbf{C}) The objects used in the latter experiment.}
    \label{fig:fig3_obj}
\end{figure}

\subsubsection{\textit{Monolithic behavioural cloning}} 

To evaluate the benefits of decomposition, we compared all four methods against a monolithic BC baseline (MT-ACT+) that uses the same BC implementation as the alignment and interaction BC policies in BC-BC, BC-Ret and Ret-BC. Instead of training separate policies for alignment and interaction, MT-ACT+ consists of a single policy trained to handle entire manipulation trajectories (Methods subsection ``Behavioural cloning implementation").

\subsection{Controlled experiments}\label{sec:controlled}

To evaluate how performance scales with different data regimes, we designed two complementary experiments that independently varied dataset size (number of demonstrations per task) and diversity (number of tasks). In the first experiment, we studied how methods scale with more demonstrations on a fixed task set. We selected four diverse micro skills spanning articulated object manipulation, deformable object interaction, scooping, and insertion. For each micro skill, we included three seen and two unseen tasks, totalling 12 seen and 8 unseen tasks (Figure~\ref{fig:fig3_obj}.A). We evaluated all methods by scaling from one to 50 demonstrations per task, where 50 demonstrations were shown to be sufficient for learning complex manipulation trajectories~\cite{Zhao2023aloha}.
 
In the second experiment, we fixed the total demonstrations at 150 and studied performance as they were distributed across more tasks. This experiment assessed whether performance degrades with fewer demonstrations per task, or if methods could benefit from more object instances despite fewer demonstrations per task. We selected 10 micro skills (Figure~\ref{fig:fig3_obj}.A and \ref{fig:fig3_obj}.B), and scaled from 10 tasks (15 demonstrations each) to 30 tasks (5 demonstrations each), and finally 50 tasks (three demonstrations each). For consistency, each diversity regime included two unseen tasks per micro skill (20 total). All objects are shown in Figure~\ref{fig:fig3_obj}.C.

During both experiments, we conducted three evaluations per task and averaged the results across all micro skills. For each evaluation, we randomised the object's position within the $80 \times 45$ cm task space, and orientation within $\pm180^\circ$ of the demonstration pose around the vertical axis.

\subsubsection{\textit{Performance overview}}

\begin{figure}[t]
    \centering
    \includegraphics[width=\linewidth]{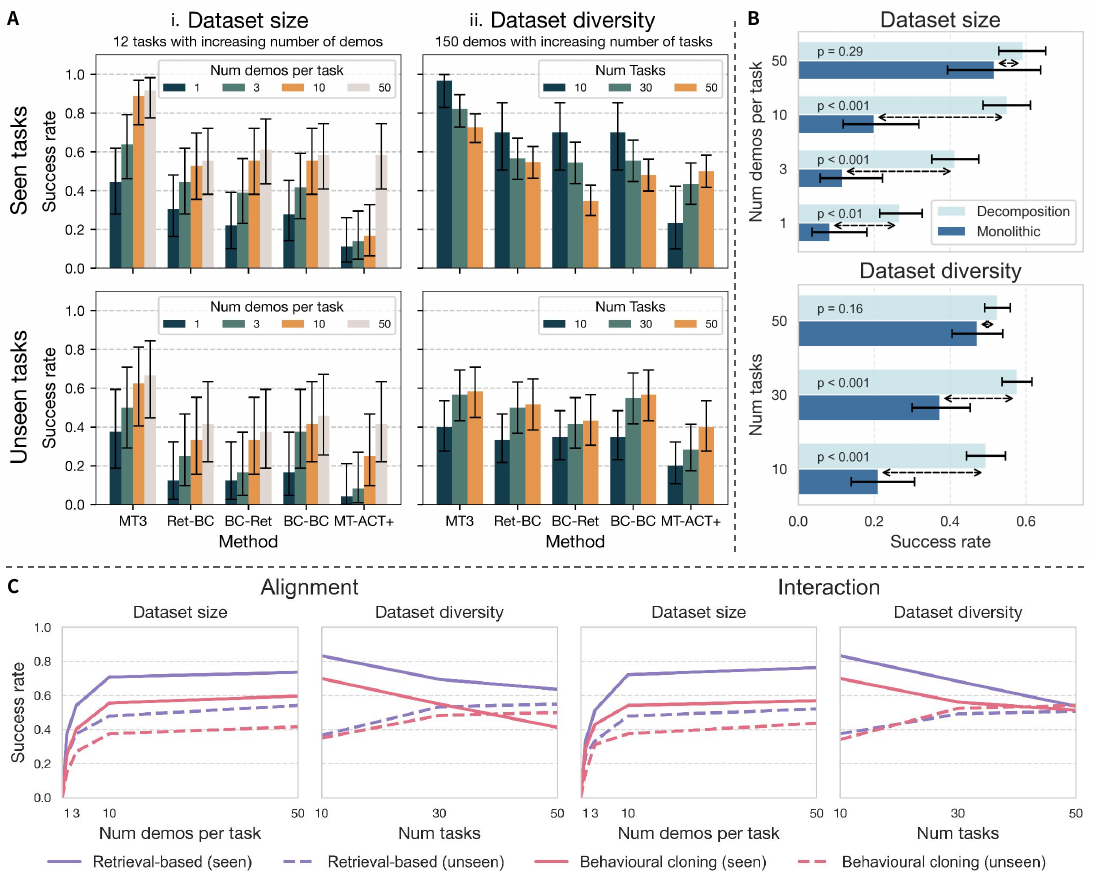}
    \caption{\textbf{Analysis of dataset size and diversity effects on task performance.} (\textbf{A}) Performance comparison across all considered methods, with error bars showing 95\% Wilson confidence intervals. For seen and unseen task sets, sample sizes were n=36 and n=24, respectively. (\textbf{B}) Comparison between decomposition-based approaches (aggregated results from Ret-Ret (MT3), Ret-BC, BC-Ret, and BC-BC) and monolithic learning (MT-ACT+), averaged across seen and unseen tasks, with error bars showing 95\% Wilson confidence intervals. Sample sizes for each comparison are detailed in Methods subsection “Statistical analysis". Statistical significance was assessed using the two-proportion Z-test. (\textbf{C}) Analysis of alignment and interaction strategies: alignment plots compare BC (BC-BC, BC-Ret) versus retrieval (Ret-BC, Ret-Ret (MT3)) for alignment, whereas interaction plots compare BC (BC-BC, Ret-BC) versus retrieval (BC-Ret, Ret-Ret (MT3)) for interaction. Success rates are shown as a function of dataset size (number of demonstrations per task) and diversity (number of tasks).}
    \label{fig:fig4_results}
\end{figure}

Figure~\ref{fig:fig4_results}.A shows the results from our dataset size and diversity experiments, revealing a clear performance hierarchy. MT3, the fully retrieval-based method, consistently demonstrated superior performance across all considered data regimes. This is particularly evident in the finding that for both seen and unseen tasks, MT3 with just three demonstrations per task outperformed all other methods, even when they were provided with 50 demonstrations per task. Moreover, the strong performance of MT3 on unseen tasks demonstrated that, despite its simplicity, retrieval is a viable approach for tackling generalisation to unseen object instances.

Decomposition also showed clear benefits, as decomposition-based methods (Ret-BC, BC-Ret, and BC-BC) generally outperformed the monolithic baseline MT-ACT+. Notably, this benefit held even when the underlying method remained unchanged, as BC-BC outperformed MT-ACT+ despite both using identical BC implementations. Figure~\ref{fig:fig4_results}.B demonstrates the advantage of the decomposition prior over a single policy by comparing the average performance of all four decomposition-based methods (Ret-Ret (MT3), Ret-BC, BC-Ret, and BC-BC) against MT-ACT+. Decomposition-based methods consistently outperformed the monolithic baseline across all data regimes tested, with the largest performance gap observed when demonstrations per task were most limited.

Finally, within the decomposition framework, Figure~\ref{fig:fig4_results}.C demonstrates that methods that used retrieval for alignment achieved higher average success rates than those that used BC (when averaged across interaction methods). Similarly, methods that used retrieval for interaction outperformed those that used BC for interaction (when averaged across alignment approaches).

\subsubsection{\textit{Scaling dataset size}}

When increasing demonstrations per task (Figure~\ref{fig:fig4_results}.A.i), all methods showed improved performance on seen and unseen tasks. This improvement stemmed from different mechanisms for each approach: retrieval-based methods benefitted from more demonstrations to select from, increasing the likelihood of finding one well-suited to the test instance and configuration. Conversely, BC methods could learn better representations and improve spatial generalisation due to greater coverage of manipulation scenarios across demonstrations.

Based on the observed trends, we would expect MT-ACT+ to eventually outperform decomposition-based methods given sufficient data, although the exact crossover point would depend on the number of tasks and their similarity. Figure~\ref{fig:fig4_results}.B reveals why this overtake is likely: decomposition-based methods achieved rapid early gains with 1–10 demonstrations per task but plateaued near 50, whereas the monolithic baseline accelerated in the 10–50 range, narrowing the performance gap.

We hypothesised that this occurred because decomposition-based methods leverage built-in task structure, enabling strong performance with limited data, but constraining their learning capacity. In contrast, the monolithic approach must learn task structure from scratch — requiring more data initially — but lacks structural constraints, allowing it to continue improving and potentially surpass decomposition-based methods with abundant data.

\subsubsection{\textit{Scaling dataset diversity}}

When we distributed a fixed demonstration budget across an increasing number of tasks (Figure~\ref{fig:fig4_results}.A.ii), we effectively added more object instances per micro skill. This diversity scaling revealed contrasting effects on retrieval-based methods: unseen task performance improved as retrieval accessed more object instances, yielding closer matches to test objects, whereas seen task performance paradoxically degraded due to an inherent trade-off in the retrieval process.

This trade-off emerged because geometry-based retrieval had to balance two competing objectives when multiple object instances exist per micro skill: selecting demonstrations with similar object pose (for successful trajectory transfer~\cite{vitiello2023one}) versus similar object geometry (for trajectory suitability for interacting with the test object). Our retrieval system considers both factors simultaneously but must make trade-offs. It may select a demonstration with better pose similarity on a different object instance, producing a trajectory less suited for the test instance. Alternatively, it may select a demonstration on the same object instance with a very different pose, making transfer harder due to large pose differences. This fundamental tension between trajectory transferability and suitability became more pronounced as object instances per micro skill increased.

The geometry-pose trade-off described above specifically affected scaling within micro skills (adding more object instances). However, retrieval-based approaches handle a different type of scaling effectively. When comparing across the two experimental paradigms while fixing demonstrations per task (light green lines in A.i versus orange lines in A.ii of Figure~\ref{fig:fig4_results}), retrieval-based methods showed either increased or stable performance despite increasing tasks from 12 to 50. This occurred because the first experiment considered four micro skills, whereas the second considered 10. The hierarchical retrieval design could explain this resilience: initial language-based filtering isolates demonstrations for the target micro skill, preventing interference when adding new macro and micro skills.

In contrast, MT-ACT+ benefitted from task diversity for both seen and unseen performance, learning more general latent representations by identifying patterns across different object instances. This benefit was substantial: learning 50 tasks across 10 micro skills with 150 demonstrations (Figure~\ref{fig:fig4_results}.A.ii) achieved comparable performance to learning 12 tasks across four micro skills with 600 demonstrations (Figure~\ref{fig:fig4_results}.A.i), effectively trading diversity for data efficiency.

However, decomposition prevented this beneficial representation sharing — BC-BC did not show the same diversity benefits as MT-ACT+, despite using identical BC implementations. We hypothesise this occurred because effective BC learning requires diverse data with shared structural patterns. Monolithic BC learned from complete trajectories that naturally combined consistent alignment patterns with diverse interaction patterns, providing the optimal balance for representation learning. Decomposition disrupted this by isolating alignment learning on synthetic linear trajectories and interaction learning on sparse but diverse real demonstrations (Methods subsection ``Behavioural cloning implementation"), preventing the synergistic learning that monolithic approaches achieved. A further discussion of the effect of decomposition on the Pareto front of performance relative to dataset diversity and learning efficiency can be found in the Appendix~\ref{app:Pareto}, and a discussion about spatial, visual appearance, and category-level generalisation of the different methods can be found in Appendix~\ref{app:Comprehensive-generalisation-overview}.

\subsection{Evaluating MT3 capabilities and failure modes across one thousand tasks}

\begin{figure}[t]
    \centering
    \includegraphics[width=0.78\linewidth]{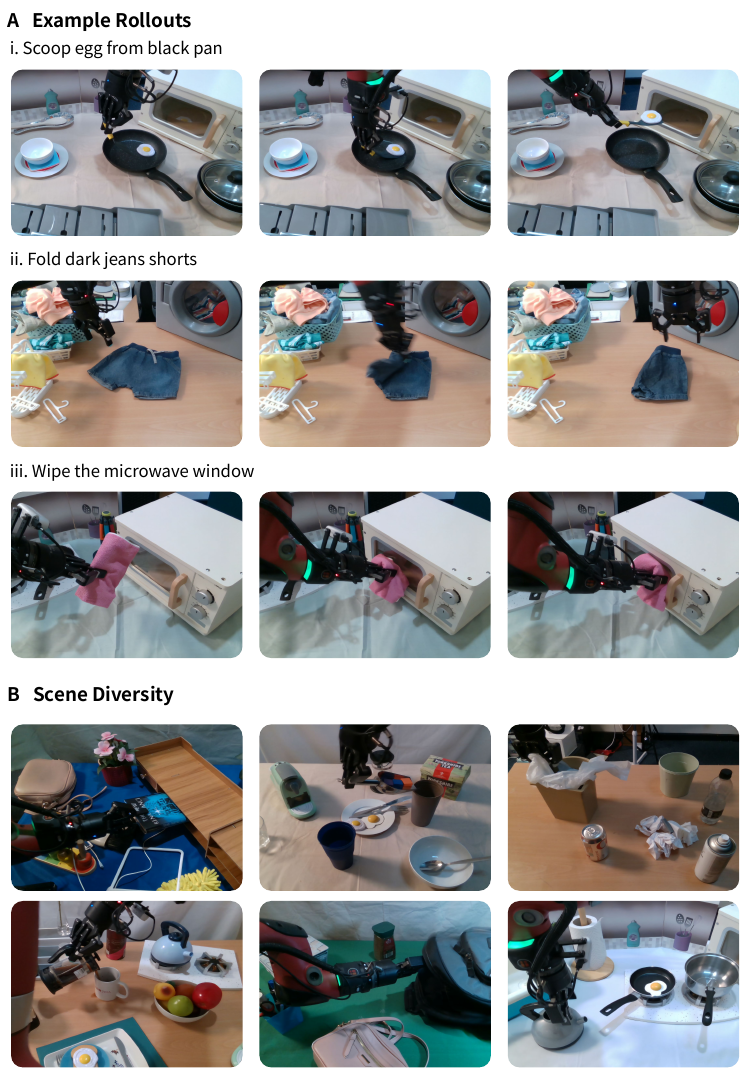}
    \caption{\textbf{Example rollouts and scene diversity from the one thousand tasks evaluation.} (\textbf{A}) Examples of recorded rollouts from the 1,000 task experiment. (\textbf{B}) Examples of the scene diversity to which MT3 was subject to during evaluation.}
    \vspace{-1em}
    \label{fig:fig5_rollouts}
\end{figure}

Although numerous studies have explored scaling monolithic BC across both tasks and demonstrations using large numbers of demonstrations per task~\cite{bharadhwaj2023roboagent, Brohan2023rt1, rt2, zhao2024aloha}, far less attention has been paid to scaling BC alternatives, especially in the minimal demonstration-per-task regime. Our controlled experiments demonstrated that MT3 was highly effective when per-task demonstrations were limited, but a crucial question remained: what are the practical boundaries of retrieval-based decomposition methods when applied to diverse real-world manipulation tasks at scale?

To investigate this, we conducted a large-scale evaluation in which we taught a robot 1,000 distinct manipulation tasks from a single demonstration each. This experiment served three analytical purposes. Firstly, we identified the specific task characteristics where MT3 excels versus where it struggles. Secondly, we sought to understand the fundamental limitations of open-loop replay for interaction. Thirdly, we aimed to characterise when global-geometry-based retrieval and cross-instance pose estimation succeed versus fail for category-level generalisation.

\subsubsection{\textit{Performance overview}}

\begin{wrapfigure}{r}{0.55\textwidth}
    \centering
    \vspace{-3em}
    \includegraphics[width=\linewidth]{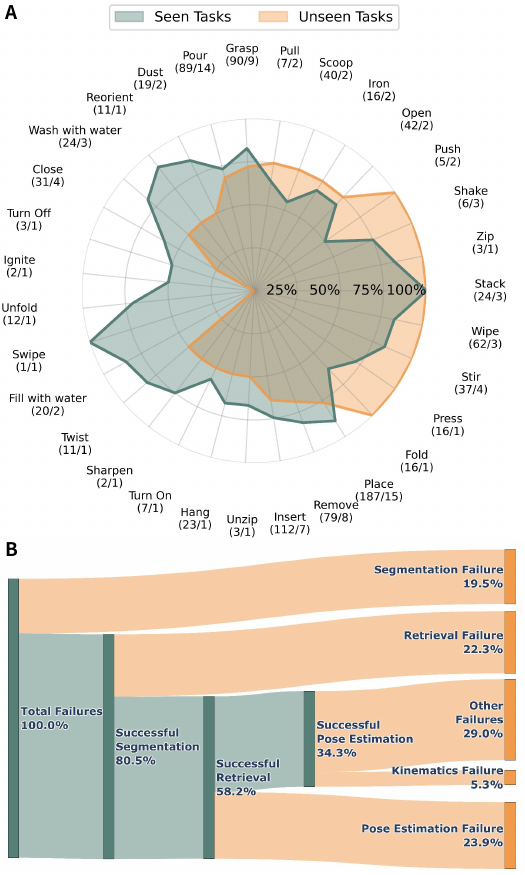}
    \caption{\textbf{Results on one thousand tasks.} (\textbf{A}) Performance of MT3 on 1,000 different seen tasks and 100 unseen ones. The results are aggregated by macro skill. (\textbf{B}) Analysis of the different failure causes experienced throughout the 1,000 tasks experiment.}
    \vspace{-1em}
    \label{fig:fig6_1000tasks}
\end{wrapfigure}

Our evaluation spanned 31 macro skills and 534 micro skills (detailed in Appendix~\ref{app:seen-tasks}), involving 402 different objects. We deliberately included tasks spanning a spectrum of complexity: from tasks well-suited to open-loop execution, including stacking, to those potentially requiring closed-loop control, such as manipulating deformable objects, and from tasks where success depends on global geometry alignment, for example, placing mugs on plates, to those requiring precise alignment of small geometric features, including inserting a coin into a slot of a piggy bank. To evaluate generalisation, we tested on an additional 100 unseen tasks spanning the same macro skills (detailed in Appendix~\ref{app:unseen-tasks}). All demonstrations were collected on a single robot over 17 hours.

Our evaluation consisted of 2,200 total rollouts (two trials per task for both seen and unseen categories — see Figure~\ref{fig:fig5_rollouts} for example rollouts) across challenging real-world conditions. These conditions included 5-20 distractor objects, varied lighting, randomised placement with up to $\pm45^\circ$ rotation, all designed to stress-test the effectiveness of MT3. We compiled these real-world rollouts into a dataset and made them open-source for the community.

Figure~\ref{fig:fig6_1000tasks}.A shows MT3's performance across different macro skills, with the numbers below each macro skill name indicating the count of seen and unseen tasks evaluated. Due to practical constraints, the distribution of tasks across macro skills is not uniform, with some macro skills having smaller task counts, leading to more variability between seen and unseen task performance. MT3 achieved a 78.25\% average success rate on seen tasks and 68\% on unseen tasks. Beyond examining the aggregate averages, we analysed the more meaningful patterns that emerged when comparing success rates in relation to task characteristics.

\subsubsection{\textit{Spatial generalisation}}

For tasks that permitted small deviations in approach angles and contact positions, such as wiping, stirring, placing, and grasping, the policy consistently achieved success rates exceeding 80\%. Open-loop replay worked effectively for such tasks since the interaction phase can accommodate minor positioning inaccuracies while rarely affecting task outcomes.

However, even for some high-tolerance tasks, MT3 encountered difficulties when small geometric features broke object symmetry and were critical for task success. This limitation stemmed from our pose estimator, which predominantly focused on global geometry, occasionally misregistering these small yet task-critical features. For instance, when grasping a kettle, its spout, being small relative to the main body, can be overlooked, leading to a $180^\circ$ orientation error and subsequent task failure.

Beyond tasks where small features broke symmetry, insertion and operations requiring high-precision alignment also proved challenging. For example, inserting a plug into a socket or hanging small keys require millimetre-level precision alignment. Since the interaction is open-loop, the interaction policy could not compensate for small alignment errors, making the system particularly vulnerable to pose estimation errors for high-precision tasks.

\subsubsection{\textit{Category-level generalisation}}

Tasks with interaction trajectories that remained consistent across object instances succeeded in generalisation. Wiping motions transferred reliably between different surfaces, and placing objects followed similar patterns regardless of minor geometric variations. For example, ``grasp mug" succeeded because the handle location relative to the body was approximately consistent across different mug designs.

However, failures occurred when changes in object instance geometry caused large variations in the required interaction trajectories. For example, pouring from a kettle required aligning the spout with the edge of a receptacle, and changing the receptacle's geometry might have required the robot to slightly adapt the pouring motion. Similarly, the swiping task failed because although MT3 could match the overall cash register geometry, it could not specifically align to the instance-variant card slot that was central to task success.

Another limiting factor arose from our retrieval approach, which fundamentally could not interpolate between demonstrated behaviours. In general, when the required trajectory lies between two demonstrated trajectories, retrieval will select one of the demonstrated ones rather than generating an appropriate intermediate solution. This binary selection process prevented adaptation to object instances that required trajectories not explicitly demonstrated.

Finally, tasks involving deformable objects proved particularly challenging because visual similarity alone was insufficient to infer the required trajectory. Different instances of deformable objects have distinct dynamic properties — such as stiffness and elasticity — that are not apparent from visual observation but critically affect manipulation success. For example, inserting a book into a backpack often failed because it required lifting the backpack flap with the book, and this dynamic interaction varied substantially across different backpacks despite visual similarity.

\subsubsection{\textit{Limitations of open-loop interaction}}

As demonstrated by our results, MT3 could proficiently tackle a huge variety of tasks. Nonetheless, our large-scale evaluation exposed fundamental limitations of open-loop trajectory replay. This approach often failed for tasks requiring online adaptation during execution. Once a trajectory begins, there is no mechanism for the policy to detect errors or adjust a course mid-execution. Although it might be possible to detect task failure and retry, this approach is often suboptimal. For instance, tasks involving deformable objects, such as folding fabric, demanded continuous adjustments based on how the material responded to manipulation.  More fundamentally, this open-loop approach cannot satisfy the requirement for reactive control. Operations like reorienting objects through contact or multi-step pushing would require continuous feedback and adaptation - capabilities that open-loop replay inherently cannot provide.

\subsubsection{\textit{Systematic failure analysis}}

To provide insight into the most common failure modes, we analysed failure cases on seen tasks, with an expert evaluator assessing each rollout for correct segmentation, exact retrieval, pose estimation success, and motion execution (Figure~\ref{fig:fig6_1000tasks}.B).

Retrieval emerged as the primary challenge (22.3\% of failures) with failures occurring most frequently with partially occluded objects or when relevant geometric variations involved small object parts that our global matching approach could not reliably identify. Segmentation and pose estimation contributed 19.5\% and 23.9\% of failures, respectively. Although segmentation issues with transparent objects and cluttered scenes should diminish with advancing models, pose estimation challenges with dramatic pose changes revealed fundamental limitations when object asymmetries create substantially different partial point clouds. The remaining 29\% of failures were predominantly from tasks with grasped objects (20.2\%), where inconsistent grasp poses between demonstration and deployment highlighted another systematic limitation of the open-loop approach - any deviation in initial conditions can propagate throughout trajectory execution without possibility for correction. However, despite its limitations, MT3 demonstrated effectiveness across many practical manipulation tasks. Our evaluation of 1,000 diverse tasks showed that many everyday tasks - from grasping and placement to insertion and washing - can be learned efficiently from single demonstrations.

\section{Discussion}

\subsection{Learning from limited per-task demonstrations}

Our controlled experiments demonstrated that decomposition and retrieval-based policies excelled in the low demonstration-per-task regime, with MT3 consistently outperforming all alternatives. This effectiveness stemmed from specialised policies that exploited distinct inductive biases rather than learning complex mappings from limited data. For alignment, explicit pose estimation and motion planning addressed spatial generalisation through analytical geometric reasoning. This bypassed the need to learn spatial relationships from limited data — a requirement that BC methods had. Similarly, for interaction, trajectory replay ensured sensible interaction execution by directly preserving demonstrated motion patterns, bypassing the challenging problem of learning complex and precise manipulation dynamics from sparse examples. These analytical biases enabled retrieval-based methods to achieve strong performance with minimal data while maintaining predictable behaviour.

\begin{wrapfigure}{r}{0.45\textwidth}
    \centering
    \vspace{-1.5em}
    \includegraphics[width=\linewidth]{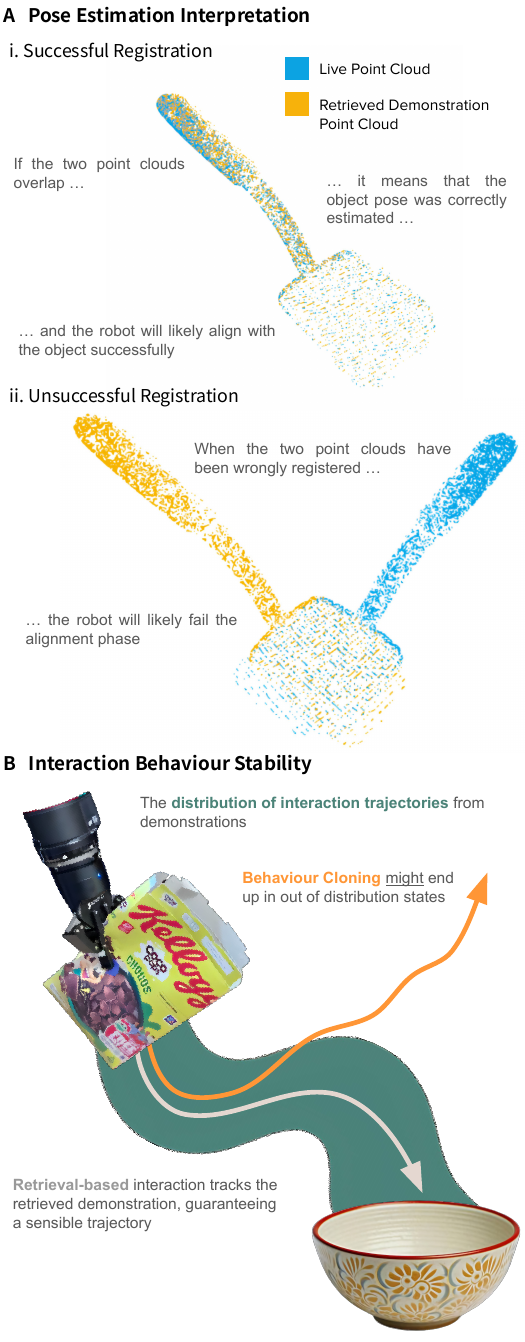}
    \caption{\textbf{Interpretability and stability of retrieval-based methods.} (\textbf{A}) Demonstration of the easily interpretable pose estimation component used by the retrieval-based alignment. (\textbf{B}) Visualisation of the stability analysis regarding using BC and retrieval for interaction.}
    \vspace{-3.5em}
    \label{fig:fig7_stability}
\end{wrapfigure}

Beyond performance advantages in low demonstration-per-task regimes, MT3 offered practical benefits through its inherent interpretability and streamlined new task acquisition. For alignment, users could visualise pose estimation results by overlapping registered point clouds (Figure~\ref{fig:fig7_stability}.A), enabling pre-emptive execution halting when misregistration was detected. For interaction, the system directly tracked demonstrated trajectories retrieved from the data buffer, ensuring behaviour never deviated from what was shown during demonstrations (Figure~\ref{fig:fig7_stability}.B). This interpretability provides clear advantages over BC methods, where policy decisions remain opaque, abstracted away within neural network weights. Furthermore, incorporating new tasks required only appending demonstrations to the existing dataset, avoiding fine-tuning and retraining procedures that BC methods typically require. This difference makes retrieval-based approaches particularly suitable for applications demanding frequent acquisition of new tasks.

\subsection{Data requirements and scaling properties of behavioural cloning}

BC's poor performance in low demonstration-per-task regimes stemmed from the simultaneous learning challenges it faced when data was scarce. BC had to concurrently learn object geometry understanding, spatial reasoning, and control from limited data while conditioning on language descriptions to distinguish between tasks. In our experimental setting, BC policies had to generalise across spatial variations ($\pm180^\circ$ rotations and varied positions across a large workspace) and geometric differences between object instances within categories. This setting proved particularly challenging for BC when data was scarce, as the policy struggled to identify meaningful patterns across sparse examples and instead resorted to memorising the few demonstrations provided, which severely limited generalisation performance.

Specifically, BC was forced to learn how to map point clouds to trajectory re-orientations and control actions, tasks that MT3 handled analytically through a pose estimator and motion planning for alignment, and trajectory replay for interaction. This analytical approach completely bypassed the need to learn these mappings from limited demonstrations.

Nevertheless, BC's scaling properties appeared more promising as demonstrations increased. Although MT3's inductive biases proved helpful in low-data regimes, they limited scaling potential as retrieval methods select single demonstrations and prevent knowledge sharing across demonstrations and tasks. In contrast, monolithic BC could exploit shared patterns across complete trajectories and benefitted from increased demonstration diversity, likely making it a more suitable approach than MT3 when data collection resources are unconstrained.

\subsection{Limitations and scope of the current study}

Our study focused on single-interaction, single-arm manipulation tasks with consistent grasped-object poses between demonstration and deployment. For all evaluated methods, varying grasp poses could be overcome without additional demonstrations using existing approaches~\cite{papagiannis2024adapting}, since all methods use segmented target object point clouds as the input. We did not evaluate environments where distractors interfered with demonstrated trajectories. Open-loop replay would have ignored obstacles and likely caused collisions, whereas BC policies would have encountered out-of-distribution scenarios requiring additional demonstration data. As such, these environmental constraints would have been expected to decrease success rates for methods reliant on retrieval-based interaction and increase data requirements for BC methods and were deemed beyond the scope of the paper.

Our reliance on vision alone made it impossible to infer dynamic properties of deformable objects without tactile feedback. Although BC could adapt through closed-loop control, open-loop retrieval-based interaction often failed since it treated novel objects exactly as training instances and committed to fixed trajectories regardless of material differences. Our approach also relied on accurate object segmentation, which can fail with transparent objects or cluttered scenes, although improving segmentation models would mitigate these issues over time. Extension to bimanual manipulation would require dual-arm coordination mechanisms, though the alignment-interaction decomposition has already been demonstrated for bimanual tasks~\cite{wang2025one-shot}, suggesting MT3's principles could extend to dual-arm scenarios. For multi-stage tasks, our single-interaction primitives could be chained using high-level planning and skill chaining~\cite{ichter2022do, vemprala2024chatgpt, Wang2024surveyllm, liang2023codeaspolicies, argus2024compositional}. Further discussion of MT3 limitations can be found in the Appendix~\ref{app:MT3-limitations}.

Despite these limitations, our findings demonstrated that decomposition combined with retrieval offered a compelling alternative to monolithic behavioural cloning when demonstration data was limited. The choice between approaches should depend on application constraints — decomposition excels at rapid task acquisition from minimal data, whilst monolithic BC becomes preferable when data collection resources are less constrained and data more diverse. Finally, our large-scale evaluation identified key technical challenges for retrieval-based approaches. Addressing these challenges whilst preserving data efficiency advantages represents an important direction for future work.

\section{Materials and methods}

\subsection{Demonstration data collection and processing}

We denote a demonstration
\begin{equation}
    \tau = \{ o_i, e_i \}_{i=1}^N
\end{equation}
as sequences of observations $o$ and end-effector states $e$ recorded at 30 Hz, where $i$ indexes time-steps and $N$ is the sequence length. Each observation $o_i$ is an RGB-D image from a calibrated head-mounted camera. The corresponding end-effector state $e_i$ includes the 6D pose of the end-effector frame $E$ in the robot's base frame $R$, $\pose{RE} \in SE(3)$, and the binary gripper state that indicates whether the gripper is open or closed. Each demonstration was paired with a language description $l$ to differentiate between tasks. This created a dataset $D$ of $M$ demonstrations and their corresponding descriptions: 
\begin{equation}
    D=\{\tau_j, l_j\}_{j=1}^M.
\end{equation}

During data collection, we recorded only the interaction phase of each task. This approach was motivated by the distinct requirements of each phase. Alignment requires achieving a specific end-effector pose relative to the target object, whereas the exact trajectory path is less critical. In contrast, interaction requires precise trajectory execution that captures task-relevant manipulation dynamics.

In practice, the demonstrator began recording demonstrations when the end-effector reached a pose suitable for initiating the intended interaction. The alignment target pose was extracted as the first pose of the recorded interaction trajectory. This selective recording strategy enabled synthetic generation of alignment trajectories when needed for BC training (Methods subsection ``Simulating alignment trajectories for behavioural cloning").
 
After data collection, we segmented all RGB-D images and converted them to target object point clouds. Segmentation enhanced efficiency and robustness of all methods against background changes and distractors. For the initial RGB image of each demonstration, we used Grounding DINO~\cite{Liu2025groundingdino} to segment the target object using the object name extracted from the task description $l$ via template-based natural language parsing~\cite{gunawardena2010automatic}. More sophisticated approaches using large language models could be employed for more complex task descriptions. For subsequent frames, we propagated the target object segmentation using XMem~\cite{cheng2022xmem}, which handled partial and full occlusions.
 
We converted segmented RGB-D images to target object point clouds using known camera parameters, with coordinate frame representation depending on the method. Retrieval-based methods used point clouds expressed in the robot frame to enable both geometry- and pose-based retrieval and to support pose estimation for alignment (Methods subsection ``Retrieval-based alignment and interaction"). BC policies used point clouds expressed in the end-effector frame to improve learning efficiency and spatial generalisation~\cite{liu2022frame}. For all demonstrations, we precomputed the geometry embedding of the target object point cloud from the first frame to enable efficient retrieval at test time.

\subsection{Retrieval-based alignment and interaction}\label{app:retrieval}


\subsubsection{\textit{Hierarchical retrieval}}

\begin{figure}[t]
    \centering
    \vspace{-1em}
    \includegraphics[width=0.9\linewidth]{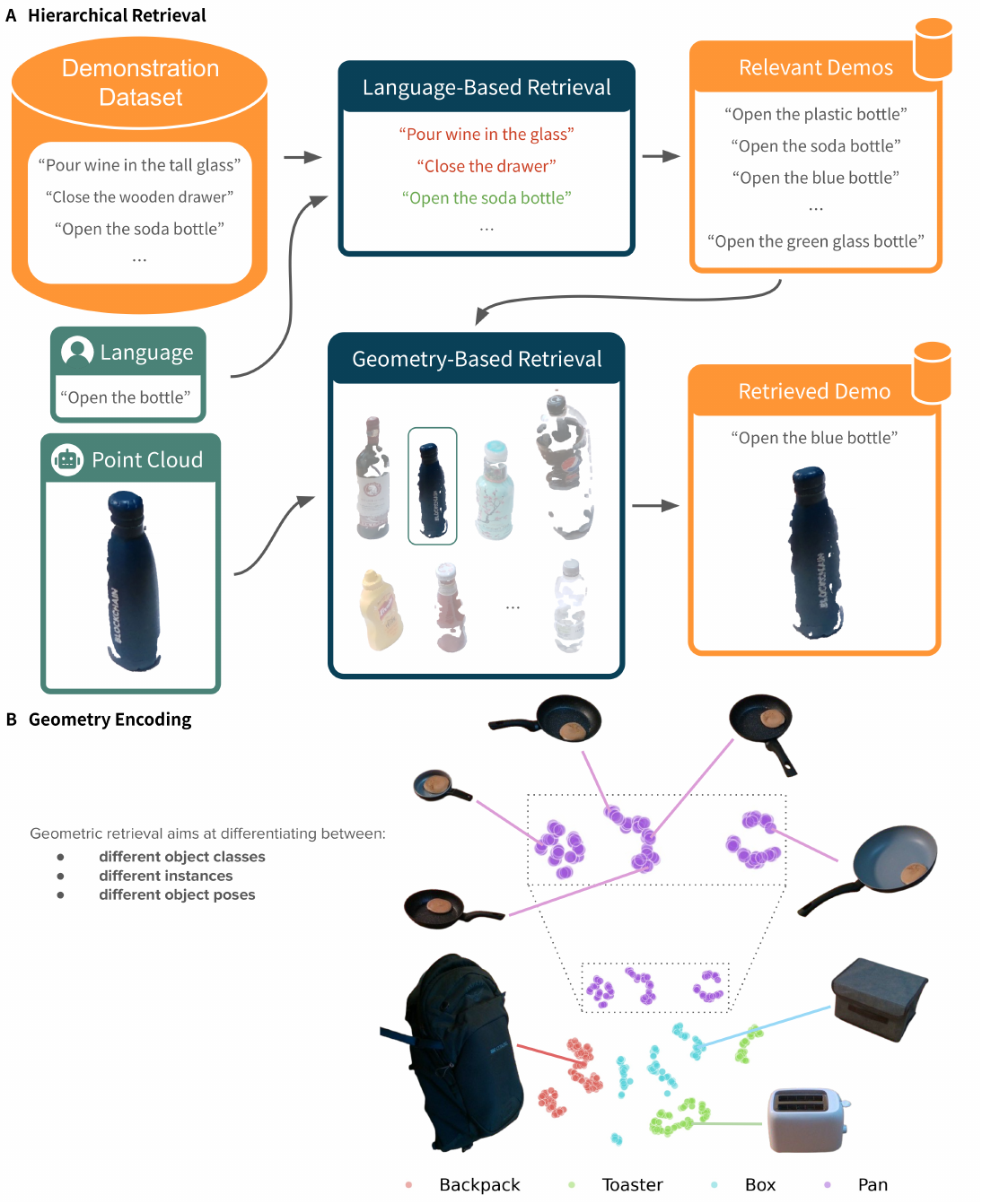}
    \caption{\textbf{Language and geometry-based retrieval.} (\textbf{A}) Hierarchical Retrieval Pipeline: Language-based retrieval identifies all demonstrations corresponding to the described micro skill. This is followed by geometry-based retrieval, which matches object shape and pose to select the single most relevant demonstration. (\textbf{B}) t-SNE visualisation of geometry encodings from the dataset size experiment with 50 demonstrations per object, showing clustering by object category (backpack, toaster, box, and pan). Each category exhibits subclusters corresponding to different object instances, with similar geometries positioned closer in the embedding space (box, toaster). Within subclusters, points from similar object poses have closer embeddings.}
    \vspace{-1em}
    \label{fig:fig8_retrieval}
\end{figure}

Our retrieval system uses a two-stage approach illustrated in Figure~\ref{fig:fig8_retrieval}.A. Firstly, we extracted the micro skill name from the task description $l$ using template matching and identified all demonstrations for the same micro skill, for example, ``open bottle". This language-based filtering served a crucial architectural role by separating task specification from execution. This design offloaded a portion of the task complexity from the policy, allowing it to focus solely on replicating the behaviour of the specific demonstration provided as context. Although we used template-based extraction and matching for simplicity, more sophisticated approaches using large language models could be employed for complex task descriptions (see Appendix~\ref{app:llm-retrieval}).

Secondly, we selected the demonstration with the most similar object to the test object in terms of pose and geometry. Geometric similarity ensures that objects requiring similar interactions are matched, whereas pose similarity minimises covariate shift for the pose estimator used by the alignment policy. To capture geometry and pose similarity, we used a PointNet++~\cite{Qi2017PointNet++} based encoder trained to predict occupancy grids using the dataset from~\cite{vitiello2023one}. The demonstration with highest cosine similarity to the test object embedding was selected.

Figure~\ref{fig:fig8_retrieval}.B shows a t-SNE plot of object embeddings from our controlled experiment with 50 demonstrations per task, revealing clustering by object category with subclusters for different instances. This hierarchical organisation enables effective generalisation: similar global geometries cluster together to match manipulation requirements, whereas pose clustering within subclusters enables relevant demonstration selection for novel object configurations.

\subsubsection{\textit{Retrieval-based alignment}}

At inference, the retrieval-based alignment policy receives the target object point cloud and a demonstration of the desired task, and its goal is to align the end-effector and the target object in the same way as shown at the beginning of the demonstration. To this end, the policy firstly used geometric reasoning to infer the required end-effector pose for the test scene and then reached this pose through motion planning.

In this work, we calculated the end-effector pose for the test scene that aligns the end-effector and target object in the same way as shown at the beginning of the demonstration using Trajectory Transfer~\cite{vitiello2023one}. The intuition behind Trajectory Transfer is that given the relative target object pose between the demonstration and test scene $\pose{\delta}$, we can map the end-effector pose at the beginning of the demonstration to the test scene using
\begin{equation}
    \pose{WE}^{Test} = \pose{\delta}\pose{WE}^{Demo}
\end{equation}

where $\pose{WE}^{Test}$ and $\pose{WE}^{Demo}$ are the end-effector poses for the test and demonstration scenes, respectively, that correspond to the same end-effector to target object pose. We estimated $\pose{\delta}$ by refining the output of the regression method proposed by~\cite{vitiello2023one} using the Open3D~\cite{Zhou2018Open3D} implementation of Generalised ICP~\cite{Segal2009}.

\subsubsection{\textit{Retrieval-based interaction}}

Similar to the retrieval-based alignment policy, at inference, the interaction policy received a demonstration of the desired task. The demonstrated trajectory was then replicated in the test scene by executing the demonstrated end-effector velocities expressed in the end effector frame.

\subsection{Behavioural cloning implementation}\label{app:bc}

We used the same network architecture and loss function to learn to align and interact with objects, and to learn the single policy for the MT-ACT+ baseline. The only difference between these applications was the training data they relied on. Below we describe our chosen backbone architecture, the loss function used, and the data, all these policies have been trained on. 

\subsubsection{\textit{Network architecture and design choices}}

Our policy architecture was required to address three key requirements across all applications. Firstly, it had to process point cloud and language inputs to enable fair comparison with retrieval-based components. Secondly, it had to effectively handle multi-task learning to support evaluation across diverse tasks. Thirdly, it had to capture the multi-modal nature of manipulation demonstrations, where multiple valid trajectories may exist for completing the same task or task phase.
 
To handle point cloud inputs, we employed a PointNet++~\cite{Qi2017PointNet++} which used the Contrastive Language-Image Pre-training (CLIP)~\cite{Radford2021LearningTV} embedding of the task description $l$ to adapt point cloud features for specific tasks using Feature-wise Linear Modulation (FiLM)~\cite{perez2018film}. To address the multi-modal nature of demonstrations, we employed variational inference, which enabled the policy to model the multi-modal distribution of valid actions. Although diffusion models offer an alternative approach, we chose variational inference as it has demonstrated strong performance for multi-task imitation learning from limited demonstrations~\cite{bharadhwaj2023roboagent}.

We adapted the MT-ACT~\cite{bharadhwaj2023roboagent} architecture to meet these requirements, modifying it to handle point cloud inputs. Additional differences from MT-ACT include incorporating action history as input to help infer task progress, removing proprioception from the input which our preliminary experiments showed improved spatial generalisation, and adding a terminal action output to explicitly signal task completion. We refer to our backbone architecture as MT-ACT+. To ensure peak performance under all experimental conditions, we independently optimised per data regime the number of network parameters for each method that used a BC policy.

\subsubsection{\textit{Loss function}}

Similar to the network architecture, the loss function used to train all BC policies was kept consistent. During training, all policies maximised the log-likelihood of demonstration action chunks, that is, 
\begin{equation}
    \underset{\theta}{\min}\text{ } \Sigma_{o_i, a_i, l\sim D} \text{ }\pi_{\theta} \left(a_{i:i+k}| o_i, l \right),
\end{equation}
with the standard VAE objective which has a reconstruction loss and a term that regularises the encoder to a Gaussian prior. Here, $o_i$ and $a_{i:i+k}$ are a sampled target object point cloud and an action chunk (Methods subsection ``Additional demonstration processing"), with $i$ representing the timestep, $k$ being the action chunk horizon, and $l$ the corresponding task description. We further augmented this loss by using learned weighting with homoscedastic uncertainty~\cite{Kendall2017Geometric} to automatically learn the weighting between different components of the reconstruction loss. 
 


 
\subsubsection{\textit{Common data augmentation steps}}

We applied common data augmentation steps during BC training regardless of whether the policy learned alignment, interaction, or both phases. To improve robustness to partial occlusions and varied object poses, we randomly masked portions of the target object point cloud using furthest point sampling followed by nearest neighbour clustering to create $10$ clusters, of which we randomly masked $4$. We also added Gaussian noise to both point clouds and action history labels to improve robustness to sensor noise.

For interaction policies specifically, we applied additional augmentation to improve robustness to covariate shift when learning from limited data. During training, we perturbed the end-effector pose within $0.9$ cm and $5$ degrees of its original position and orientation, then updated the corresponding state and action labels to reflect this perturbation. This augmentation helped the policy handle small deviations from demonstrated trajectories that could have occurred during deployment and was enabled by using target object point clouds expressed in the end-effector frame. Additional details regarding data processing specific to BC interaction policies can be found in the Appendix~\ref{app:Additional-demonstration-processing}.

\subsubsection{\textit{Simulating alignment trajectories for behavioural cloning}}

To train alignment capabilities, both the alignment BC policies (used by BC-Ret and BC-BC) and the MT-ACT+ baseline required trajectories that reached the initial pose of each demonstration. We simulated 1,000 alignment trajectories per demonstration by sampling starting poses within a $30\times80\times80$ cm cuboid above the robot's taskspace and generating linear trajectories to the demonstrated initial end-effector pose.

Since our BC policies used target object point clouds expressed in the end-effector frame, we could generate training data by reusing the same target object point cloud across different virtual end-effector poses. Specifically, we took the target object point cloud from the first demonstration frame and used it at every waypoint along simulated linear trajectories that end at the target alignment pose. This generated synthetic observation-action pairs where each waypoint had the same geometry of the point cloud input but different action outputs corresponding to the remaining trajectory to reach the target pose. We generated waypoints with $1$ cm spacing along each linear path.

To improve alignment accuracy, we supplemented the training data with additional observation-action pairs near the final alignment pose. For each waypoint in a simulated alignment trajectory, we generated an additional observation-action pair by perturbing the end-effector pose within $1$mm-$1$ cm and $0.5$-$5$ degrees of the target alignment pose.
 
\subsubsection{\textit{Combining simulated alignment trajectories and demonstrations}}

Our monolithic baseline, MT-ACT+, required training data for both the alignment and interaction phases of demonstrated tasks. As such, we combined the simulated alignment trajectories with demonstrated trajectories to create a dataset of entire manipulation trajectories, adjusting the history and action labels at the boundary between alignment and interaction phases.

\subsubsection{\textit{Statistical analysis}}

Experimental outcomes were obtained from $n$ independent repeats for each considered task. This data is represented in Figures~\ref{fig:fig4_results} and \ref{fig:fig6_1000tasks} as mean values calculated across all associated tasks and repeats. Error bars in Figures~\ref{fig:fig4_results}.A and \ref{fig:fig4_results}.B represent 95\% Wilson confidence intervals. The two-proportion Z-test was applied to assess the statistical significance of performance differences between all considered decomposition-based methods and the monolithic baseline, with p-values shown in Figure~\ref{fig:fig4_results}.B.

For Figures \ref{fig:fig4_results}.A and  \ref{fig:fig4_results}.B, the number of evaluated trajectories depended on the experiment configuration. For Figure \ref{fig:fig4_results}.A dataset size experiment, the sample size was $n=36$ for seen tasks and $n=24$ for unseen tasks. For Figure  \ref{fig:fig4_results}.A dataset diversity experiment, when learning 10 tasks, $n=30$ (seen) and $n=60$. When learning 30 tasks, $n=90$ (seen) and $n=60$ (unseen). Finally, when learning 50 tasks, $n=150$ (seen) and $n=60$ (unseen).

For Figure \ref{fig:fig4_results}.B dataset size experiment, $n=240$ (decomposition) and $n=60$ (monolithic). For the dataset diversity experiment, when learning 10 tasks, $n=120$ (decomposition) and $n=30$ (monolithic). When learning 30 tasks, $n=200$ (decomposition) and $n=50$ (monolithic). When learning 50 tasks, $n=280$ (decomposition) and $n=70$ (monolithic).


\bibliography{bibliograph}  

\clearpage
\begin{appendices}

\section{The effect of decomposition on the diversity-efficiency Pareto front of performance}\label{app:Pareto}

\begin{figure}[h] 
	\centering
	\includegraphics[width=\linewidth]{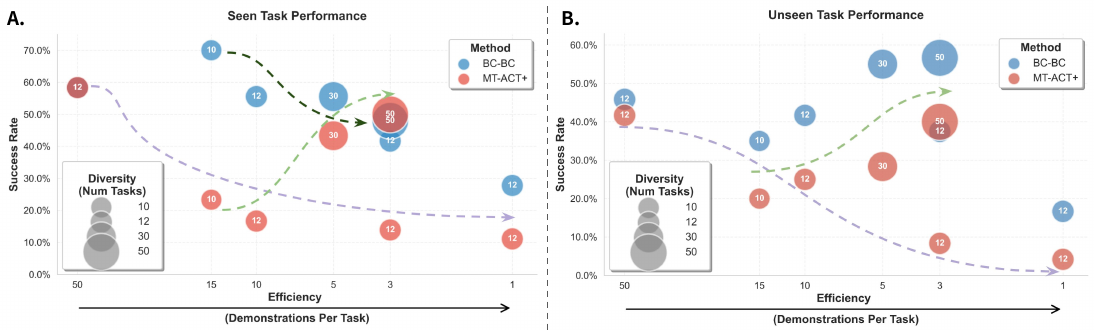} 
    \caption{\textbf{The effect of decomposition on the diversity-efficiency Pareto front of performance.} (\textbf{A}) Seen task performance comparing monolithic BC (MT-ACT+) against decomposition-based BC (BC-BC). Purple dashed line shows the traditional Pareto front when task diversity is held approximately constant, revealing that both methods exhibit similar performance-efficiency trade-offs. Light green arrow indicates monolithic BC's beneficial scaling when distributing a fixed demonstration budget across more tasks, simultaneously improving both efficiency and performance. Dark green arrow shows decomposition's negative scaling under the same conditions, where increased diversity coupled with reduced demonstrations per task degrades performance. (\textbf{B}) Unseen task performance showing similar traditional Pareto fronts (purple dashed line). Both methods benefit from increased diversity when fixing the demonstration budget (light green arrow).}	\label{fig:pareto} 
\end{figure}

We analysed how decomposition affects the Pareto front of performance relative to both learning efficiency (demonstrations per task) and dataset diversity (number of tasks) by combining results from our dataset size and diversity experiments. To isolate decomposition's effects, we compared monolithic BC (MT-ACT+) against decomposition using identical BC implementations (BC-BC). These aggregated results are shown in Figure~\ref{fig:pareto}, where the x-axis represents learning efficiency (number of demonstrations per task), the y-axis shows average success rate, and the size of each data point indicates the number of tasks being learned.

\subsection{Traditional efficiency-performance trade-offs}

For both seen (Figure~\ref{fig:pareto}.A) and unseen (Figure~\ref{fig:pareto}.B) tasks, the purple dashed lines illustrate the traditional Pareto front shape. When keeping the number of tasks approximately constant, performance decreased as the number of demonstrations per task decreased. Both monolithic BC and decomposition displayed similar front shapes, but decomposition consistently achieved higher absolute success rates across all considered numbers of demonstrations per task.

The performance gap between methods was smallest when learning from 50 demonstrations per task and increased in lower demonstrations-per-task regimes, though not monotonically. When demonstrations per task where limited, decomposition's explicit separation of trajectories into alignment and interaction phases made the learning problem more tractable, providing clear structure that would otherwise need to be discovered from data. Monolithic BC required substantially more data to learn this task structure that decomposition received by construction, explaining the larger performance gaps in data-scarce regimes.

\subsection{Scaling with diversity}

When considering performance on seen tasks, decomposition fundamentally alters the relationship between diversity (the number of object instances) and efficiency as shown by the green arrows in Figure~\ref{fig:pareto}.A. When maintaining three demonstrations per task, increasing task diversity (the number of object instances) from 12 to 50 tasks — effectively increasing the total dataset from 36 to 150 demonstrations — boosted performance for both methods. However, monolithic BC showed dramatically larger gains compared to decomposition's modest improvement.

The green arrows reveal contrasting Pareto fronts when simultaneously optimizing for both diversity and efficiency by distributing a fixed demonstration budget across an increasing number of tasks. The light green arrow shows that monolithic BC improved efficiency by distributing a fixed demonstration budget across more tasks — a highly desirable property for practical applications. In stark contrast, decomposition did not show this beneficial scaling (dark green arrow), with performance degrading as diversity increased and the number of demonstrations per task decreased.

In contrast, for unseen tasks, Figure~\ref{fig:pareto}.B reveals that both methods benefited from increased diversity (light green arrow), though decomposition maintains its absolute performance advantage.

\subsection{Mechanistic insights}

We hypothesise that this divergent scaling behaviour stems from how each approach leverages demonstration data. Monolithic BC learns from complete trajectories that naturally combine consistent alignment patterns with diverse interaction patterns, creating ideal conditions for learning shared representations that benefit both spatial and category-level generalisation. As diversity increases, the method can extract more generalisable features from the consistent alignment components while simultaneously learning geometry-dependent patterns from the varied interaction components.

Decomposition disrupts this synergistic learning by separating training across two incompatible data distributions: alignment policies train exclusively on high-volume but monotonous synthetic linear trajectories, whereas interaction policies learn from low-volume but highly diverse real demonstrations. This separation prevents the unified representation learning that enables monolithic BC's beneficial diversity scaling. The alignment phase lacks the complexity needed to benefit from increased diversity, whereas the interaction phase suffers from extreme data sparsity when demonstrations are distributed across more tasks.

The contrasting effects of diversity on seen versus unseen performance reveal how trajectory decomposition affects different types of generalisation. Geometry-dependent trajectory variations primarily manifest during the interaction phase — different object shapes require fundamentally different manipulation strategies. This enables the interaction policy to learn geometry-based generalisation, explaining improved unseen task performance as diversity increases. However, the alignment phase training data consists of synthetic linear trajectories that lack geometry-dependent variations. The alignment policy thus learns spatial reasoning that is largely geometry-agnostic. When pose coverage per object decreases (as demonstrations spread across more tasks), we hypothesize that the alignment policy may default to average poses or struggle with the reduced pose variations, explaining the degraded seen task performance where spatial generalisation is critical.

For unseen tasks, both approaches benefit because category-level generalisation primarily depends on exposure to diverse object geometries within each category. The interaction phase — where geometric differences are most pronounced — can leverage this diversity regardless of whether it operates within a monolithic or decomposed architecture. This explains the universal benefit of diversity for unseen tasks, contrasting with the architecture-dependent effects observed for seen tasks.

\clearpage

\section{Comprehensive generalisation overview}\label{app:Comprehensive-generalisation-overview}

Our systematic evaluation revealed how different methods generalise across three key dimensions when learning from few demonstrations per task: spatial generalisation, visual appearance generalisation, and category-level generalisation.

For spatial generalisation, successful alignment determines performance. We observed that methods using retrieval-based alignment outperform those using BC alignment in terms of spatial generalisation, especially in low demonstration regimes. The alignment BC policy had to learn to interpolate or extrapolate across spatial variations from limited demonstration data, frequently operating outside its training distribution. In contrast, the retrieval-based alignment policy leveraged a pose estimator trained on $\pm45^\circ$ variations in object poses, allowing it to remain within its training distribution more frequently even with limited demonstrations per task. This difference was particularly evident in precision-dependent tasks like opening boxes and insertion toasts into toasters, where BC's out-of-distribution spatial reasoning often lead to imprecise alignment and subsequent task failures.

For visual appearance generalisation, we found that all methods managed to cope well with visual appearance variations and lighting changes. Methods that used the retrieval-based alignment policy achieved robustness through geometry-only encodings for demonstration retrieval and domain randomisation during pose estimator training. Similarly, the BC alignment policy generalised through visual variations present in demonstration data. Segmentation helped both approaches by isolating the task-relevant object from backgrounds and visual distractors.

For category-level generalisation, we observed that all methods exhibit a strong correlation between generalisation success and whether interaction trajectories depend on global versus local object geometry. When trajectories where primarily determined by global geometry — such as placing spoons on spoon holders — both BC and retrieval methods succeed across object instances. However, when success depended on local geometric features, all methods struggled.

BC methods lacked sufficient data to learn which local features are task-relevant with limited per-task demonstrations. For retrieval methods, the pose estimation component operates in a task-agnostic manner, predominantly focusing on global geometry without understanding which specific geometric features are relevant for task success. We also observed failures across all methods when manipulating deformable objects like backpacks and shorts, where instance-dependent properties affect manipulation dynamics but cannot be reliably inferred from visual geometry alone.

\clearpage

\section{Distribution of our one thousand seen tasks}\label{app:seen-tasks}

\begin{longtable}{llc}
\caption{Distribution of one thousand seen tasks used in our large-scale evaluation of MT3 across macro and micro skills. Macro skills represent broad manipulation primitives, including picking up and inserting, whereas micro skills denote specialised variations requiring distinct motion profiles, such as ``pick up mug".}\label{app:table:num-tasks-per-skill}\\
\toprule
\textit{}\textbf{Macro Skill} & \textbf{Micro Skill} & \textbf{Number of Tasks} \\
\midrule
\endfirsthead
\toprule
\textbf{Macro Skill} & \textbf{Micro Skill} & \textbf{Number of Tasks} \\
\midrule
\endhead
\bottomrule
\endfoot
\bottomrule
\endlastfoot
\begin{tabular}{@{}l@{}}place object on/in/next to\\another object\end{tabular} & & 187 \\
\midrule
 & place plate on coaster & 10 \\
 & place shoe on shelf & 7 \\
 & place shoe next to shoe & 7 \\
 & place shorts in basket & 7 \\
 & place mug on coaster & 7 \\
 & place spoon on spoon rest & 6 \\
 & place kettle in sink & 6 \\
 & place bowl in sink & 6 \\
 & place plate in sink & 6 \\
 & place bottle in bin & 4 \\
 & place mug on plate & 4 \\
 & place mug in coffee machine & 4 \\
 & place notepad on shelf & 3 \\
 & place toast on plate & 3 \\
 & place phone on charger & 3 \\
 & place shirt in cabinet & 2 \\
 & place stapler in cabinet & 2 \\
 & place jeans in cabinet & 2 \\
 & place kettle on left-stove & 2 \\
 & place post-it in cabinet & 2 \\
 & place kettle on right-stove & 2 \\
 & place handbag on shelf & 2 \\
 & place glass on coaster & 2 \\
 & place marble in box & 2 \\
 & place mug in sink & 2 \\
 & place ball in cabinet & 2 \\
 & place container on shelf & 2 \\
 & place telephone in holder & 2 \\
 & place cloth in basket & 2 \\
 & place bowl on plate & 2 \\
 & place pepper in air fryer & 1 \\
 & place pepper on shelf & 1 \\
 & place phone next to desk organiser & 1 \\
 & place pasta in bag & 1 \\
 & place pickle in box & 1 \\
 & place plate in dishwasher & 1 \\
 & place pencil in holder & 1 \\
 & place pear in box & 1 \\
 & place peach in bowl & 1 \\
 & place peach in box & 1 \\
 & place rose in pot & 1 \\
 & place plate on tray & 1 \\
 & place socks in basket & 1 \\
 & place tissue on tray & 1 \\
 & place sugarbag on shelf & 1 \\
 & place strawberry in box & 1 \\
 & place stapler next to desk organiser & 1 \\
 & place spoon in spoon rest & 1 \\
 & place spoon in sink & 1 \\
 & place spoon in orginiser & 1 \\
 & place shirt in suitcase & 1 \\
 & place postit next to desk organiser & 1 \\
 & place shirt in basket & 1 \\
 & place scissors in holder & 1 \\
 & place salt on shelf & 1 \\
 & place rose in vase & 1 \\
 & place paper in bin & 1 \\
 & place queen on chessboard & 1 \\
 & place pot in dishwasher & 1 \\
 & place paracetamol in basket & 1 \\
 & place aglaomorpha in pot & 1 \\
 & place pan on right-stove & 1 \\
 & place can on shelf & 1 \\
 & place fork in orginiser & 1 \\
 & place fork in dishwasher & 1 \\
 & place flowers in vase & 1 \\
 & place cranesbill in pot & 1 \\
 & place coaster on mousepad & 1 \\
 & place cloth in suitcase & 1 \\
 & place carrot in air fryer & 1 \\
 & place capsule in washing machine & 1 \\
 & place bowl on tray & 1 \\
 & place fork on tray & 1 \\
 & place bowl in dishwasher & 1 \\
 & place bottle on shelf & 1 \\
 & place bottle in suitcase & 1 \\
 & place book next to desk organiser & 1 \\
 & place block on jenga tower & 1 \\
 & place bib in basket & 1 \\
 & place banana in bowl & 1 \\
 & place apple in box & 1 \\
 & place fork in sink & 1 \\
 & place glass on shelf & 1 \\
 & place pan on left-stove & 1 \\
 & place knife on tray & 1 \\
 & place paintroll on holder & 1 \\
 & place orange in bowl & 1 \\
 & place notepad next to desk organiser & 1 \\
 & place apple in bowl & 1 \\
 & place mug in dishwasher & 1 \\
 & place mouse on mousepad & 1 \\
 & place lemon in bowl & 1 \\
 & place leaf in cocktail shaker & 1 \\
 & place knife in orginiser & 1 \\
 & place grass in pot & 1 \\
 & place keyboard on mousepad & 1 \\
 & place jeans in suitcase & 1 \\
 & place jeans in basket & 1 \\
 & place impatiens in pot & 1 \\
 & place ibuprofen in basket & 1 \\
 & place hole puncher next to desk organiser & 1 \\
 & place holder next to desk organiser & 1 \\
 & place handbag on desk organiser & 1 \\
 & place vase on shelf & 1 \\
\midrule
\begin{tabular}{@{}l@{}}insert object into\\another object\end{tabular} & & 112 \\
\midrule
 & insert shorts into washing machine & 11 \\
 & insert plate into microwave & 10 \\
 & insert bowl into microwave & 10 \\
 & insert teabag into mug & 10 \\
 & insert straw into glass & 9 \\
 & insert toast into toaster & 7 \\
 & insert plate into dish rack & 6 \\
 & insert brush into bottle & 5 \\
 & insert book into backpack & 4 \\
 & insert towel into washing machine & 4 \\
 & insert bottle into backpack & 4 \\
 & insert paper into tray & 3 \\
 & insert ring into peg & 3 \\
 & insert toothbrush into holder & 2 \\
 & insert paper roll into holder & 2 \\
 & insert socks into washing machine & 2 \\
 & insert trapezoid into shape organiser & 1 \\
 & insert star into grab board & 1 \\
 & insert square into shape organiser & 1 \\
 & insert triangle into grab board & 1 \\
 & insert square into grab board & 1 \\
 & insert spout into bottle & 1 \\
 & insert plug into outlet & 1 \\
 & insert shirt into washing machine & 1 \\
 & insert rectangle into grab board & 1 \\
 & insert book into desk organiser & 1 \\
 & insert pen into holder & 1 \\
 & insert pen into desk organiser & 1 \\
 & insert paper into hole puncher & 1 \\
 & insert muddler into cocktail shaker & 1 \\
 & insert lego into lego & 1 \\
 & insert ice clip into cocktail shaker & 1 \\
 & insert cork into bottle & 1 \\
 & insert coin into piggy bank & 1 \\
 & insert circle into grab board & 1 \\
 & insert usb into computer & 1 \\
\midrule
pick up object & & 90 \\
\midrule
 & pick up glass & 9 \\
 & pick up kettle & 8 \\
 & pick up shoe & 7 \\
 & pick up bowl & 6 \\
 & pick up lid & 5 \\
 & pick up mug & 5 \\
 & pick up bottle & 5 \\
 & pick up can & 3 \\
 & pick up box & 3 \\
 & pick up bag & 3 \\
 & pick up trash & 3 \\
 & pick up toothbrush from glass & 2 \\
 & pick up sponge & 2 \\
 & pick up plum & 1 \\
 & pick up plush & 1 \\
 & pick up rose & 1 \\
 & pick up pickle & 1 \\
 & pick up strawberry & 1 \\
 & pick up spoon & 1 \\
 & pick up stapler & 1 \\
 & pick up peach & 1 \\
 & pick up t-shirt & 1 \\
 & pick up teabag from teabox & 1 \\
 & pick up telephone from holder & 1 \\
 & pick up pear & 1 \\
 & pick up apple & 1 \\
 & pick up paper & 1 \\
 & pick up paintroll from holder & 1 \\
 & pick up orange & 1 \\
 & pick up mouse & 1 \\
 & pick up lemon & 1 \\
 & pick up knife & 1 \\
 & pick up jug & 1 \\
 & pick up jeans & 1 \\
 & pick up iron & 1 \\
 & pick up hole puncher & 1 \\
 & pick up handbag & 1 \\
 & pick up fork & 1 \\
 & pick up eraser & 1 \\
 & pick up cloth & 1 \\
 & pick up banana & 1 \\
 & pick up trousers & 1 \\
\midrule
\begin{tabular}{@{}l@{}}pour from container\\into another container\end{tabular} & & 89 \\
\midrule
 & pour from coffee press into mug & 10 \\
 & pour from scoop into glass & 9 \\
 & pour from mixer into glass & 9 \\
 & pour from jigger into glass & 9 \\
 & pour from pan into plate & 9 \\
 & pour from carton into bowl & 6 \\
 & pour from spoon into mug & 6 \\
 & pour from cereal box into bowl & 4 \\
 & pour from cereal box into pot & 4 \\
 & pour from jug into pot & 4 \\
 & pour from kettle into mug & 4 \\
 & pour from carton into mug & 2 \\
 & pour from bottle into jigger & 2 \\
 & pour from kettle into glass & 2 \\
 & pour from bottle into pan & 2 \\
 & pour from bottle into glass & 2 \\
 & pour from watering can into aglaomorpha & 1 \\
 & pour from jigger into mixer & 1 \\
 & pour from jug into glass & 1 \\
 & pour from carton into jug & 1 \\
 & pour from watering can into impatiens & 1 \\
\midrule
\begin{tabular}{@{}l@{}}remove object from\\another object\end{tabular} & & 79 \\
\midrule
 & remove bowl from microwave & 10 \\
 & remove mug from mug rack & 10 \\
 & remove trash bag from bin & 8 \\
 & remove toast from toaster & 7 \\
 & remove toy from box & 6 \\
 & remove trousers from washing machine & 4 \\
 & remove dish from dish rack & 3 \\
 & remove hanger from mug rack & 3 \\
 & remove ring from peg & 3 \\
 & remove cup from mug rack & 2 \\
 & remove trousers from hanger & 2 \\
 & remove shirt from washing machine & 2 \\
 & remove shorts from washing machine & 2 \\
 & remove towel from washing machine & 2 \\
 & remove towel from hanger & 2 \\
 & remove keys from mug rack & 1 \\
 & remove circle from grabboard & 1 \\
 & remove triangle from grabboard & 1 \\
 & remove cable from charger & 1 \\
 & remove cap from mug rack & 1 \\
 & remove star from grabboard & 1 \\
 & remove tie from mug rack & 1 \\
 & remove mask from mug rack & 1 \\
 & remove groceries from bag & 1 \\
 & remove bulb from socket & 1 \\
 & remove rectangle from grabboard & 1 \\
 & remove plug from socket & 1 \\
 & remove square from grabboard & 1 \\
\midrule
\begin{tabular}{@{}l@{}}wipe object with\\another object\end{tabular} & & 62 \\
\midrule
 & wipe window with towel & 5 \\
 & wipe bowl with towel & 5 \\
 & wipe plate with dishmop & 5 \\
 & wipe toaster with towel & 4 \\
 & wipe shelf with dishmop & 3 \\
 & wipe sink with towel & 3 \\
 & wipe sink with sponge & 3 \\
 & wipe sink with dishmop & 3 \\
 & wipe shelf with towel & 3 \\
 & wipe bowl with sponge & 3 \\
 & wipe shelf with sponge & 3 \\
 & wipe plate with towel & 2 \\
 & wipe plate with sponge & 2 \\
 & wipe tray with dishmop & 2 \\
 & wipe tray with sponge & 2 \\
 & wipe tray with towel & 2 \\
 & wipe bowl with dishmop & 2 \\
 & wipe lunchbox with towel & 1 \\
 & wipe lunchbox with sponge & 1 \\
 & wipe lunchbox with dishmop & 1 \\
 & wipe desk organiser with towel & 1 \\
 & wipe desk organiser with sponge & 1 \\
 & wipe desk organiser with dishmop & 1 \\
 & wipe white with sponge & 1 \\
 & wipe whiteboard with eraser & 1 \\
 & wipe whiteboard with sponge & 1 \\
 & wipe whiteboard with towel & 1 \\
\midrule
open object & & 42 \\
\midrule
 & open box & 10 \\
 & open drawer & 6 \\
 & open lunch box compartment & 4 \\
 & open pot & 4 \\
 & open washing machine & 3 \\
 & open bin & 2 \\
 & open bottle & 2 \\
 & open kettle & 2 \\
 & open microwave & 2 \\
 & open cash register & 1 \\
 & open dishwasher & 1 \\
 & open mixer & 1 \\
 & open sharpener & 1 \\
 & open soap drawer & 1 \\
 & open vegetable cutter & 1 \\
 & open watering can & 1 \\
\midrule
\begin{tabular}{@{}l@{}}scoop\\from container\end{tabular} & & 40 \\
\midrule
 & scoop broccoli from pan & 9 \\
 & scoop egg from pan & 9 \\
 & scoop pancake from pan & 9 \\
 & scoop shrimp from pan & 9 \\
 & scoop beanbag & 1 \\
 & scoop coffee from tin & 1 \\
 & scoop ice from bowl & 1 \\
 & scoop sugar from bowl & 1 \\
\midrule
stir in container & & 37 \\
\midrule
 & stir spoon in mug & 10 \\
 & stir spoon in glass & 9 \\
 & stir spoon in pan & 9 \\
 & stir spoon in pot & 8 \\
 & stir spoon in cocktail shaker & 1 \\
\midrule
close object & & 31 \\
\midrule
 & close drawer & 6 \\
 & close envelope & 4 \\
 & close pot & 4 \\
 & close bottle & 2 \\
 & close box & 2 \\
 & close kettle & 2 \\
 & close microwave & 2 \\
 & close washing machine & 2 \\
 & close dishwasher & 1 \\
 & close machine & 1 \\
 & close mixer & 1 \\
 & close sharpener & 1 \\
 & close soap drawer & 1 \\
 & close vegetable cutter & 1 \\
 & close watering can & 1 \\
\midrule
stack objects & & 24 \\
\midrule
 & stack bowls & 3 \\
 & stack holders & 3 \\
 & stack mugs & 3 \\
 & stack pans & 3 \\
 & stack plates & 3 \\
 & stack glasss & 2 \\
 & stack coasters & 1 \\
 & stack cubes & 1 \\
 & stack cups & 1 \\
 & stack towels & 1 \\
 & stack trays & 1 \\
 & stack trousers & 1 \\
 & stack weights & 1 \\
\midrule
wash object with water & & 24 \\
\midrule
 & wash plate with water & 6 \\
 & wash mug with water & 4 \\
 & wash glass with water & 3 \\
 & wash bowl with water & 2 \\
 & wash colander with water & 1 \\
 & wash fork with water & 1 \\
 & wash knife with water & 1 \\
 & wash lid with water & 1 \\
 & wash pan with water & 1 \\
 & wash pot with water & 1 \\
 & wash spatula with water & 1 \\
 & wash sponge with water & 1 \\
 & wash spoon with water & 1 \\
\midrule
\begin{tabular}{@{}l@{}}hang object on\\another object\end{tabular} & & 23 \\
\midrule
 & hang mug on hanger & 10 \\
 & hang cloth on hanger & 2 \\
 & hang cup on hanger & 2 \\
 & hang cap on hanger & 1 \\
 & hang hanger on dish rack & 1 \\
 & hang hanger on hanger & 1 \\
 & hang hanger on trays & 1 \\
 & hang jeans on hanger & 1 \\
 & hang keys on hanger & 1 \\
 & hang mask on hanger & 1 \\
 & hang tie on hanger & 1 \\
 & hang trousers on hanger & 1 \\
\midrule
fill object with water & & 20 \\
\midrule
 & fill kettle with water & 7 \\
 & fill bottle with water & 6 \\
 & fill glass with water & 3 \\
 & fill can with water & 2 \\
 & fill iron with water & 1 \\
 & fill press with water & 1 \\
\midrule
\begin{tabular}{@{}l@{}}dust object with\\another object\end{tabular} & & 19 \\
\midrule
 & dust shelf with brush & 6 \\
 & dust tray with brush & 5 \\
 & dust shelf with duster & 3 \\
 & dust desk organiser with brush & 2 \\
 & dust desk organiser with duster & 1 \\
 & dust drawers with brush & 1 \\
 & dust drawers with duster & 1 \\
\midrule
fold object & & 16 \\
\midrule
 & fold shorts & 10 \\
 & fold shirt & 2 \\
 & fold trousers & 2 \\
 & fold cloth & 1 \\
 & fold coaster & 1 \\
\midrule
iron object & & 16 \\
\midrule
 & iron shorts & 10 \\
 & iron shirt & 2 \\
 & iron towel & 2 \\
 & iron trousers & 2 \\
\midrule
press object & & 16 \\
\midrule
 & press button & 6 \\
 & press keyboard key & 3 \\
 & press coffee press & 1 \\
 & press enter & 1 \\
 & press hole puncher & 1 \\
 & press pencil & 1 \\
 & press soap & 1 \\
 & press stapler & 1 \\
 & press vegetable cutter & 1 \\
\midrule
unfold object & & 12 \\
\midrule
 & unfold shorts & 10 \\
 & unfold cloth & 1 \\
 & unfold coaster & 1 \\
\midrule
reorient object & & 11 \\
\midrule
 & reorient knob & 4 \\
 & reorient sink & 4 \\
 & reorient handle & 2 \\
 & reorient lamp & 1 \\
\midrule
twist object & & 11 \\
\midrule
 & twist knob & 8 \\
 & twist lightbulb & 2 \\
 & twist lemon & 1 \\
\midrule
turn on device & & 7 \\
\midrule
 & turn on lamp & 2 \\
 & turn on toaster & 2 \\
 & turn on clock & 1 \\
 & turn on machine & 1 \\
 & turn on sink & 1 \\
\midrule
pull object & & 7 \\
\midrule
 & pull shelf & 3 \\
 & pull tray & 2 \\
 & pull napkin & 1 \\
 & pull post-it & 1 \\
\midrule
shake object & & 6 \\
\midrule
 & shake bottle & 6 \\
\midrule
push object & & 5 \\
\midrule
 & push shelf & 3 \\
 & push tray & 2 \\
\midrule
turn off device & & 3 \\
\midrule
 & turn off lamp & 2 \\
 & turn off sink & 1 \\
\midrule
unzip object & & 3 \\
\midrule
 & unzip case & 1 \\
 & unzip handbag & 1 \\
 & unzip suitcase & 1 \\
\midrule
zip object & & 3 \\
\midrule
 & zip case & 1 \\
 & zip handbag & 1 \\
 & zip suitcase & 1 \\
\midrule
light object with lighter & & 2 \\
\midrule
 & light candle with lighter & 2 \\
\midrule
\begin{tabular}{@{}l@{}}sharpen object with\\another object\end{tabular} & & 2 \\
\midrule
 & sharpen knife with sharpener & 1 \\
 & sharpen pencil with sharpener & 1 \\
\midrule
swipe on object & & 1 \\
\midrule
 & swipe card on cash register & 1 \\
\bottomrule
\end{longtable}

\clearpage

\section{Distribution of our one hundred unseen tasks}\label{app:unseen-tasks}

\begin{longtable}{llc}
\caption{Distribution of one hundred unseen tasks used in our large-scale experiment across macro and micro skills. These tasks covered the same macro skills as the seen tasks but involved novel object instances within known categories, enabling assessment of MT3's category-level generalisation capabilities.}\label{app:table:num-unseen-tasks-per-skill}\\
\toprule
\textbf{Macro Skill} & \textbf{Micro Skill} & \textbf{Number of Tasks} \\
\midrule
\endfirsthead
\toprule
\textbf{Macro Skill} & \textbf{Micro Skill} & \textbf{Number of Tasks} \\
\midrule
\endhead
\bottomrule
\endfoot
\bottomrule
\endlastfoot
\begin{tabular}{@{}l@{}}place object on/in/next to\\another object\end{tabular} & & 15 \\
\midrule
 & place kettle in sink & 4 \\
 & place plate in sink & 3 \\
 & place apple in bowl & 1 \\
 & place bowl on plate & 1 \\
 & place fork in sink & 1 \\
 & place lemon in bowl & 1 \\
 & place mug in sink & 1 \\
 & place peach in bowl & 1 \\
 & place tissue on tray & 1 \\
 & place toast on plate & 1 \\
\midrule
\begin{tabular}{@{}l@{}}pour from container\\into another container\end{tabular} & & 14 \\
\midrule
 & pour from pan into plate & 3 \\
 & pour from jug into glass & 2 \\
 & pour from jug into pot & 2 \\
 & pour from kettle into mug & 2 \\
 & pour from spoon into mug & 2 \\
 & pour from bottle into glass & 1 \\
 & pour from cereal box into pot & 1 \\
 & pour from scoop into glass & 1 \\
\midrule
pick up object & & 9 \\
\midrule
 & pick up bottle & 3 \\
 & pick up mug & 3 \\
 & pick up cloth & 1 \\
 & pick up handbag & 1 \\
 & pick up toothbrush from glass & 1 \\
\midrule
\begin{tabular}{@{}l@{}}remove object from\\another object\end{tabular} & & 8 \\
\midrule
 & remove cap from mug rack & 2 \\
 & remove shorts from washing machine & 2 \\
 & remove bowl from microwave & 1 \\
 & remove cable from charger & 1 \\
 & remove dish from dish rack & 1 \\
 & remove mug from mug rack & 1 \\
\midrule
\begin{tabular}{@{}l@{}}insert object into\\another object\end{tabular} & & 7 \\
\midrule
 & insert teabag into mug & 4 \\
 & insert bowl into microwave & 2 \\
 & insert toothbrush into holder & 1 \\
\midrule
close object & & 4 \\
\midrule
 & close envelope & 4 \\
\midrule
stir in container & & 4 \\
\midrule
 & stir spoon in mug & 4 \\
\midrule
\begin{tabular}{@{}l@{}}wipe object with\\another object\end{tabular} & & 3 \\
\midrule
 & wipe bowl with dishmop & 1 \\
 & wipe plate with dishmop & 1 \\
 & wipe plate with towel & 1 \\
\midrule
shake object & & 3 \\
\midrule
 & shake bottle & 3 \\
\midrule
stack objects & & 3 \\
\midrule
 & stack bowls & 1 \\
 & stack pans & 1 \\
 & stack weights & 1 \\
\midrule
wash object with water & & 3 \\
\midrule
 & wash pan with water & 2 \\
 & wash glass with water & 1 \\
\midrule
pull object & & 2 \\
\midrule
 & pull napkin & 1 \\
 & pull shelf & 1 \\
\midrule
iron object & & 2 \\
\midrule
 & iron towel & 2 \\
\midrule
\begin{tabular}{@{}l@{}}dust object with\\another object\end{tabular} & & 2 \\
\midrule
 & dust shelf with duster & 1 \\
 & dust tray with brush & 1 \\
\midrule
fill object with water & & 2 \\
\midrule
 & fill glass with water & 2 \\
\midrule
\begin{tabular}{@{}l@{}}scoop\\from container\end{tabular} & & 2 \\
\midrule
 & scoop egg from pan & 1 \\
 & scoop sugar from bowl & 1 \\
\midrule
push object & & 2 \\
\midrule
 & push shelf & 2 \\
\midrule
open object & & 2 \\
\midrule
 & open bin & 1 \\
 & open pot & 1 \\
\midrule
twist object & & 1 \\
\midrule
 & twist lemon & 1 \\
\midrule
turn off device & & 1 \\
\midrule
 & turn off lamp & 1 \\
\midrule
\begin{tabular}{@{}l@{}}sharpen object with\\another object\end{tabular} & & 1 \\
\midrule
 & sharpen knife with sharpener & 1 \\
\midrule
\begin{tabular}{@{}l@{}}hang object on\\another object\end{tabular} & & 1 \\
\midrule
 & hang hanger on trays & 1 \\
\midrule
unzip object & & 1 \\
\midrule
 & unzip handbag & 1 \\
\midrule
zip object & & 1 \\
\midrule
 & zip handbag & 1 \\
\midrule
unfold object & & 1 \\
\midrule
 & unfold cloth & 1 \\
\midrule
turn on device & & 1 \\
\midrule
 & turn on lamp & 1 \\
\midrule
reorient object & & 1 \\
\midrule
 & reorient lamp & 1 \\
\midrule
press object & & 1 \\
\midrule
 & press button & 1 \\
\midrule
light object with lighter & & 1 \\
\midrule
 & light candle with lighter & 1 \\
\midrule
fold object & & 1 \\
\midrule
 & fold cloth & 1 \\
\midrule
swipe on object & & 1 \\
\midrule
 & swipe card on cash register & 1 \\
\bottomrule
\end{longtable}

\clearpage

\section{MT3's limitations and future work}\label{app:MT3-limitations}

Although depth cameras struggle with transparent and reflective objects, we observed that the appearance of these objects was consistent despite not reflecting their true geometry, enabling successful manipulation even in these challenging cases. Future work could explore RGB-based unseen object pose estimation solutions to further improve robustness in these scenarios.

Regarding scalability, storage requirements for demonstration data are minimal, requiring only a single RGB-D image and a sequence of vectors (that is, poses) per demonstration. This efficient representation enables scaling to thousands of tasks even with limited storage capacity.

Finally, when the interaction trajectory depends on the pose of multiple target objects simultaneously, the framework proposed by~\cite{heppert2024ditto} can be used to adapt the interaction trajectory based on the relative pose of both objects.

\clearpage

\section{Large language models for language-based retrieval}\label{app:llm-retrieval}

As previously suggested, large language models (LLMs) could be used for the language-retrieval part of hierarchical retrieval (Methods subsection ``Retrieval-based alignment and interaction" in the main paper), which is responsible for selecting demonstrations corresponding to the test-time language command. LLMs have recently displayed in-context learning capabilities~\cite{mirchandani2023large}, which could be helpful for demonstration retrieval.

In particular, this utility extends to two key areas. Firstly, enabling the retrieval of demonstrations even when users employ highly natural language commands. Secondly, identifying the correct sequence of learned tasks required to accomplish desired long-horizon tasks.

To investigate these applications, we conducted a small proof-of-concept experiment. We instructed Grok 3, a state-of-the-art LLM, on the task of retrieval, providing it with a CSV file listing our taught micro skills. The LLM was then prompted with various language commands, simulating informal and natural user language commands.

The following prompt was used to instruct the LLM:

\begin{quotation}
\noindent\textit{You are a robotic agent. I attached here a table with the names of all the skills I have taught you (the same skill could have been repeated multiple times, but that should not matter). From here on after, I will request you to do a task. I will mark my request with "R:". You will then provide me with an answer "A:". Your answer should be one of the following:}

\textit{- If the requested task could be interpreted as one of the skills I have taught you and that is in the table I provided then your answer should be the exact name of that skill as it is written in the table.}

\textit{- In other occasions my request could be a more long-horizon task. In that case you should return a list of skills, separated by commas, which together would probably complete the request. Every skill in the list should be returned with exactly the same name as it is in the table.}

\textit{- Lastly, if my request is likely not achievable with the skills you have been taught you should answer with "none".}\\

\noindent\textit{I will now give you examples of how our interactions should look like:}

\noindent\textit{\textbf{R}: take out my trousers from the laundry machine \\
\textbf{A}: remove trousers from washing machine}\\

\noindent\textit{\textbf{R}: make tea \\
\textbf{A}: pick up teabag from teabox, insert teabag into mug, pick up kettle, pour from kettle into mug}\\

\noindent\textit{\textbf{R}: prepare me a steak \\
\textbf{A}: none}\\

\textit{You only have one arm, therefore whenever you pick something up you should place it down before interacting with another object. If no placing down skill has been taught for that object but it would be needed then return "none" as the task is unfeasible.}

\textit{When I ask about a task which is very similar to a skill I have taught you but would require an object in hand, assume you have that object in the hand. For instance in the case "R: pour me some coffee" there is a very similar skill in the table, hence you can imagine you have already grasped the coffee press and just answer with "A: pour from coffee press into mug"}
\end{quotation}

\subsection{Single-stage tasks}
After giving this set of instructions to the LLM we have fed it with a series of language commands. Hereafter, we report the conversation.

\begin{quotation}
\noindent\textit{
(\textbf{User}) R: place the spoon on its holder \\
(\textbf{LLM}) A: place spoon on spoon rest}
\end{quotation}

\begin{quotation}
\noindent\textit{
(\textbf{User}) R: the pancake is done, take it off the heat \\
(\textbf{LLM}) A: scoop pancake from pan}
\end{quotation}

\begin{quotation}
\noindent\textit{
(\textbf{User}) R: put the dish in the rack \\
(\textbf{LLM}) A: insert plate into dish rack}
\end{quotation}

\begin{quotation}
\noindent\textit{
(\textbf{User}) R: pour me some cereal in my bowl please \\
(\textbf{LLM}) A: pour from cereal box into bowl}
\end{quotation}

\begin{quotation}
\noindent\textit{
(\textbf{User}) R: clean the whiteboard with the sponge \\
(\textbf{LLM}) A: wipe whiteboard with sponge}
\end{quotation}

\begin{quotation}
\noindent\textit{
(\textbf{User}) R: rinse the glass \\
(\textbf{LLM}) A: wash glass with water}
\end{quotation}

For short-horizon, single-stage tasks, the LLM displayed proficiency in retrieving the intended micro-skills. This suggests that LLMs could indeed allow users greater flexibility in phrasing commands when interacting with the robot.

\subsection{Long-horizon tasks}

However, challenges emerged with long-horizon skill chaining, particularly when requests were vague or ill-posed. When the language command was well-defined, the LLM could reliably provide a reasonable skill sequence. For example, below we can see the skills suggested by the LLM when asked to prepare toast.

\begin{quotation}
\noindent\textit{
(\textbf{User}) R: make me some toast and plate it \\
(\textbf{LLM}) A: insert toast into toaster, remove toast from toaster, place toast on plate}
\end{quotation}

On the other hand, skill chaining can sometimes be ill-posed. For instance, when asking a robot ``Make me breakfast", unless the agent has prior knowledge of the user's preferences, the number of skill combinations that could be chained together is almost unbounded. Therefore, when asking excessively open-ended, long-horizon commands, the LLM can sometimes get overwhelmed by the possibilities and starts hallucinating incoherent task sequences. Below is an example of this behaviour.

\begin{quotation}
\noindent\textit{
(\textbf{User}) R: make me a cocktail \\
(\textbf{LLM}) A: pick up bottle, pour from bottle into jigger, pour from jigger into mixer, place leaf in cocktail shaker, insert muddler into cocktail shaker, insert ice clip into cocktail shaker, stir spoon in cocktail shaker, pour from mixer into glass}
\end{quotation}

It is clear that the LLM was able to retrieve tasks that are all related to the user's request. For example, the jigger, the shaker and the ice-clips are all cocktail making tools. Nonetheless, the relationship between these tasks lacks logic. For instance, pouring from the jigger happens without putting down the bottle first or picking up the jigger; and stirring the cocktail with the spoon naively assumes that the spoon is already there. Having said this, in the next subsections we will discuss how we believe there to be room for improvement.

\subsection{Requesting user assistance}
As it can be seen from the initial system prompt we have also instructed the LLM to acknowledge when there was no sequence of taught skills that could satisfy the user's request. We can see an example of this when we asked one of the ill-posed long-horizon tasks that we discussed earlier.

\begin{quotation}
\noindent\textit{
(\textbf{User}) R: make me a breakfast \\
(\textbf{LLM}) A: none}
\end{quotation}

Together with the above reply, the LLM also clarified that the reason for its answer was the poor definition of the requested task. Although we have not implemented it, being able to identify when skill chaining is not well defined can be used to ask the user for further information. When the LLM has an idea of what skills it could chain but recognises that such skills have not been taught, it could ask the user whether they would agree with said skill chain and then ask them to provide demonstrations for the missing skills. 

Similarly, if a request is ill-defined, the LLM could ask questions to the user to narrow down the possibilities. For instance, in the example we are considering, the LLM could ask the user if they want toast for breakfast or if they would enjoy a tea rather than a coffee. Questions like these could guide the LLM while also conveying a sense of control in the user.

\subsection{Conclusion}
Overall, we believe that current LLMs could already be used for language retrieval. In our opinion, the limitations with skill chaining could be mitigated, if not totally eliminated, by more carefully designed system prompting. Incorporating the conversation with the user, as suggested in the earlier subsection, would also likely improve performance. Moreover, these technologies are progressing very rapidly and we are especially hopeful that, in the future, language-retrieval will become a trivial task for almost any LLM.

\clearpage

\section{Additional demonstration processing}\label{app:Additional-demonstration-processing}

We further processed demonstrations specifically for BC policy training. This processing applied to interaction policies (used by Ret-BC and BC-BC) and the MT-ACT+ baseline (Methods subsection ``Combining Simulated Alignment Trajectories and Demonstrations").

Firstly, we encoded the task descriptions $l$ using CLIP~\cite{Radford2021LearningTV}. To ensure a uniform spatial resolution across demonstrations, we subsampled demonstrated trajectories to maintain a consistent 1cm distance between consecutive waypoints while preserving important events like gripper state changes. We then computed actions $a_{i:i+k}$ as relative poses between the current end-effector pose and future poses within the prediction horizon $k=3$, using the angle-axis representation for orientations. Similarly, we computed history action labels as relative poses between current and past poses within the history horizon of $10$ actions. Both parameters were manually tuned, with the choice of $k$ matching the suggestion from the authors of MT-ACT~\cite{bharadhwaj2023roboagent}.

\end{appendices}

\end{document}